\pdfoutput=1

\documentclass[11pt]{article}

\usepackage[preprint]{acl}
\usepackage{adjustbox}

\usepackage{times}
\usepackage{nameref}
\usepackage{subfigure}
\usepackage[graphicx]{realboxes}
\usepackage{latexsym}
\usepackage{multirow}
\usepackage{colortbl}
\usepackage{CJKutf8}
\usepackage{subfigure}
\usepackage[T1]{fontenc}

\usepackage[utf8]{inputenc}

\usepackage{authblk}
\usepackage{microtype}
\usepackage{hyperref}
\usepackage{inconsolata}
\usepackage{authblk}
\usepackage{graphicx}
\graphicspath{{Image/}}

\title{RAG and RAU: A Survey on Retrieval-Augmented Language Model in Natural Language Processing}

\author[1]{Yucheng Hu\thanks{Equal contribution.}}
\author[2]{Yuxing Lu\thanks{Equal contribution. Corresponding author.}}

\affil[1]{East China University of Science and Technology \\ \texttt{huyc@mail.ecust.edu.cn}}
\affil[2]{Peking University \\ \texttt{yxlu0613@gmail.com}}

\begin{document}
\maketitle
\begin{abstract}

Large Language Models (LLMs) have catalyzed significant advancements in Natural Language Processing (NLP), yet they encounter challenges such as hallucination and the need for domain-specific knowledge. To mitigate these, recent methodologies have integrated information retrieved from external resources with LLMs, substantially enhancing their performance across NLP tasks. This survey paper addresses the absence of a comprehensive overview on Retrieval-Augmented Language Models (RALMs), both Retrieval-Augmented Generation (RAG) and Retrieval-Augmented Understanding (RAU), providing an in-depth examination of their paradigm, evolution, taxonomy, and applications. The paper discusses the essential components of RALMs, including Retrievers, Language Models, and Augmentations, and how their interactions lead to diverse model structures and applications. RALMs demonstrate utility in a spectrum of tasks, from translation and dialogue systems to knowledge-intensive applications. The survey includes several evaluation methods of RALMs, emphasizing the importance of robustness, accuracy, and relevance in their assessment. It also acknowledges the limitations of RALMs, particularly in retrieval quality and computational efficiency, offering directions for future research. In conclusion, this survey aims to offer a structured insight into RALMs, their potential, and the avenues for their future development in NLP. The paper is supplemented with a Github Repository containing the surveyed works and resources for further study: \url{https://github.com/2471023025/RALM_Survey}.

\end{abstract}

\section{Introduction}
\begin{figure*}[t]
    \centering
    \includegraphics[width=\textwidth]{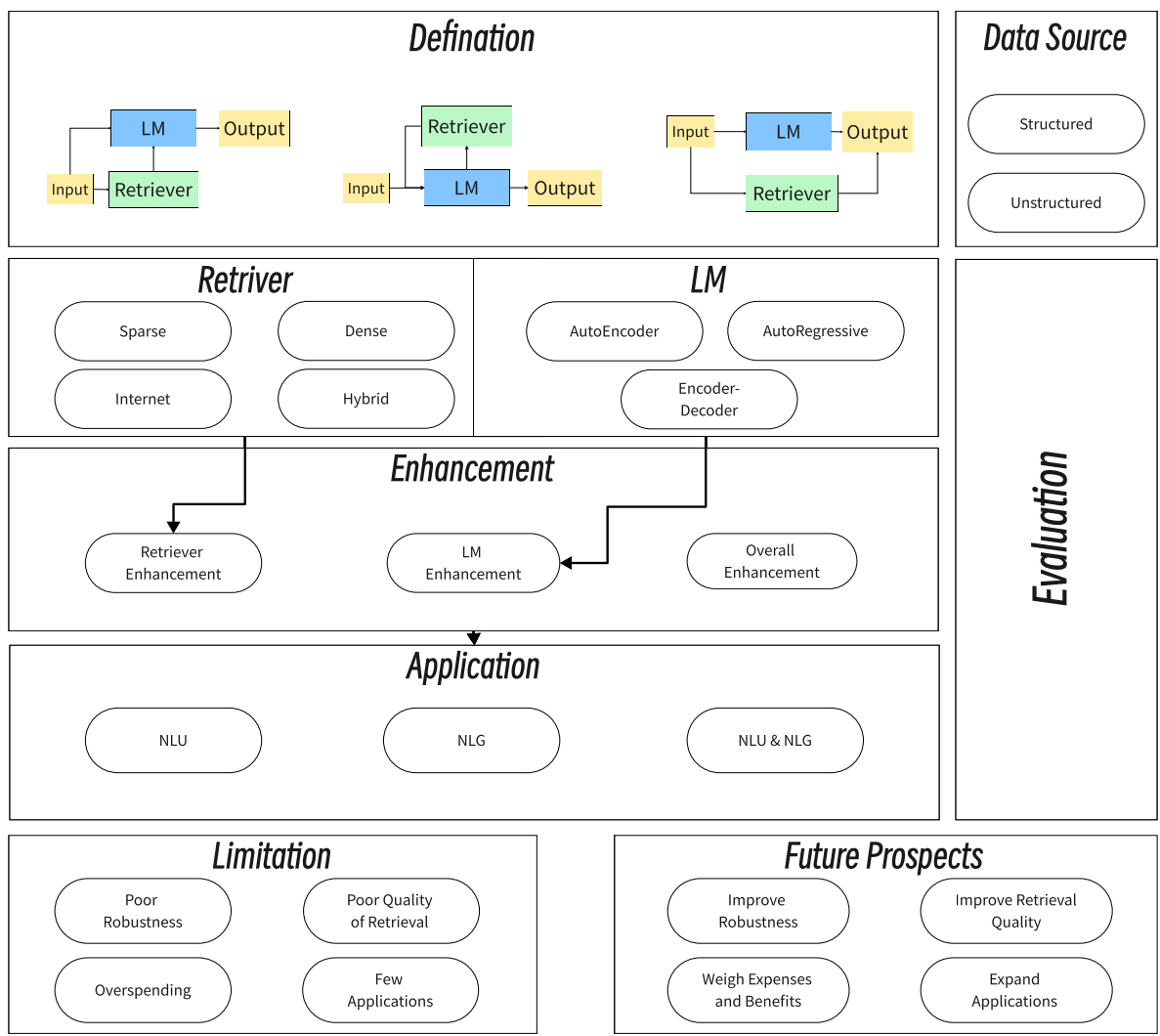}
    \caption{A general overview of this survey‘s work}
    \label{fig:overview}
\end{figure*}

Natural Language Processing (NLP) is a significant focus within the realms of computer science and artificial intelligence, dedicated to the study of theoretical and methodological frameworks that enable effective communication between humans and computers using natural language. As a multidisciplinary field, NLP integrates linguistics, computer science, and mathematics with the aim of realizing the mutual transformation between human language and computer data. Its ultimate objective is to empower computers with the capability to process and "understand" natural language, thereby facilitating tasks such as automatic translation, text categorization, and sentiment analysis. The complexity of NLP is evident in the numerous steps it encompasses, including word segmentation, part-of-speech tagging, parsing, stemming, named entity recognition, and more, all of which contribute to the challenge of replicating human language understanding in artificial intelligence systems.

Traditional natural language processing tasks typically employ statistic-based algorithms \cite{hogenboom2010overview} \cite{serra2013parnt} \cite{aussenac2005text} and deep learning algorithms such as convolutional neural network (CNN) \cite{176}, recurrent neural network (RNN) \cite{177}, long short-term memory network (LSTM) \cite{178}, and others. Recently, with the advent of the transformer architecture \cite{107} as a leading representative of natural language processing, its popularity has grown significantly. The transformer architecture, as a prominent large language model \cite{12} \cite{89} in the natural language processing domain, has consistently demonstrated enhanced performance, attracting the attention of an increasing number of researchers who are engaged in studying its capabilities.

The most prevalent LMs nowadays are the GPT families \cite{99} \cite{104} \cite{88} and Bert families \cite{90} \cite{110} \cite{135}, which have been demonstrated to excel in a multitude of natural language processing tasks. Among these, the AutoEncoder language model is particularly adept at natural language understanding tasks, while the AutoRegressive language model is more suited to natural language generation tasks. While increasing parameters \cite{86} and model tuning \cite{179} can enhance the performance of LLMs, the phenomenon of "hallucination" \cite{180} persists. Furthermore, the limitations of LMs in effectively handling knowledge-intensive work \cite{feng2023knowledge} and their inability to promptly update their knowledge \cite{mousavi2024your} are consistently apparent. Consequently, numerous researchers \cite{19} \cite{20} \cite{22} have employed the technique of retrieval to obtain external knowledge, which can assist the language model in attaining enhanced performance in a multitude of tasks.

Currently, there is a paucity of surveys on the use of retrieval augmentation to enhance the performance of LLMs. \citet{80} provide a comprehensive overview of work on RAG for multimodality. \citet{181} concentrate on the utilisation of retrieval augmentation generation techniques for the Artificial Intelligence Generated Content (AIGC) domain. This article provides a comprehensive overview of recent RAG work, but it does not cover all relevant domains. Additionally, the article lacks sufficient detail to provide a comprehensive timeline of the overall development. \citet{182} investigate the enhancement of RAG for large models. This article summarizes some of the recent RAG work, but it introduces the retrievers and generators independently, which is not conducive to the upgrading and interactions with the components of subsequent work. \citet{1} focus on text generation only. The article has fewer figures and tables, and the content is more abstract, which is not conducive to the reader's understanding.

Also, surveys on RAG only tells half of the story in retrieval-augmented methods in NLP. Not only do tasks associated with NLG require retrieval enhancement techniques, but NLU tasks also necessitate external information. To date, there is a scarcity of comprehensive surveys that thoroughly review the application of augmented retrieval techniques across the spectrum of NLP. In order to improve the current situation, this paper presents the following contributions:

\begin{figure*}[tp]
    \centering
    \includegraphics[width=\textwidth]{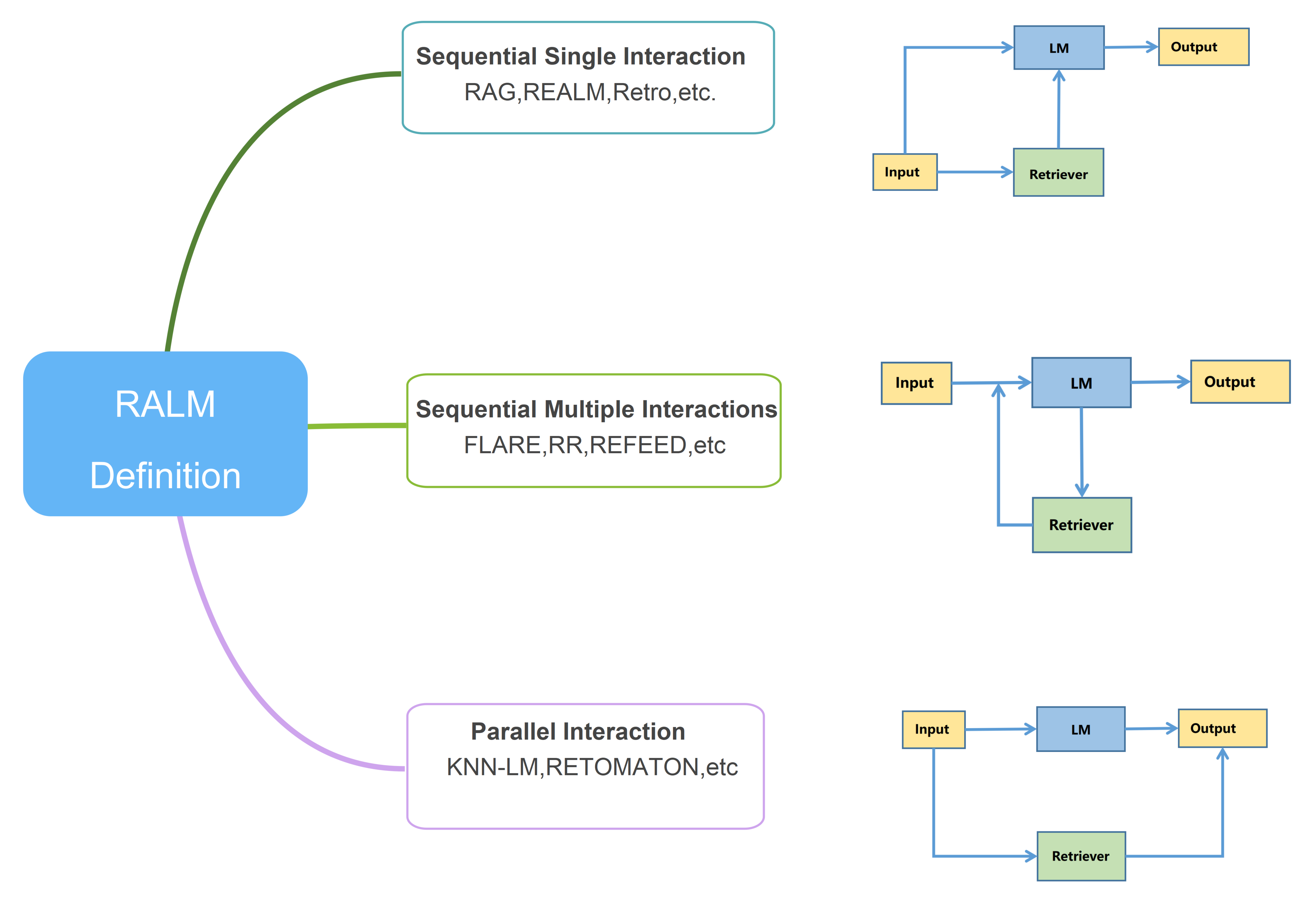}
    \caption{Three different ways the Retriever interacts with the LM}
    \label{defination}
\end{figure*}

(1) The article does not merely focus on the work related to RAG; it also places significant emphasis on RALM and aligns with the concept of NLP. The work related to generation aligns with NLG, while the rest of the work aligns with NLU.

(2) The two components of RALM, the Retriever and the Language Model, are described in detail, and the different interaction modes of these two components are precisely defined for the first time.

(3) A comprehensive overview of the RALM work schedule is provided, along with a summary of the common and novel applications of current RALM, accompanied by an analysis of the associated limitations. Potential solutions to these limitations are proposed, along with recommendations for future research directions.

Figure \ref{fig:overview} provides a general overview of the framework of RALM methods. The following is a summary of the paper:
Section \ref{s2} defines RALM.
Section \ref{s3} provides a detailed classification and summary of the work of retrievers in RALM.
Section \ref{s4} provides a detailed classification and summary of the work of LMs in RALM.
Section \ref{s5} provides a classification and summary of specific enhancements to RALM. 
Section \ref{s6} of RALM is a classification and summary of the sources of retrieved data. 
Section \ref{s7} is a summary of RALM applications. 
Section \ref{s8} is a summary of RALM evaluations and benchmarks. 
Finally, Section \ref{s9} is a discussion of the limitations of existing RALM and directions for future work.

\section{Definition}
\label{s2}

\begin{figure*}[t]
    \centering
    \vspace{-20pt}
    \includegraphics[width=0.81\linewidth]{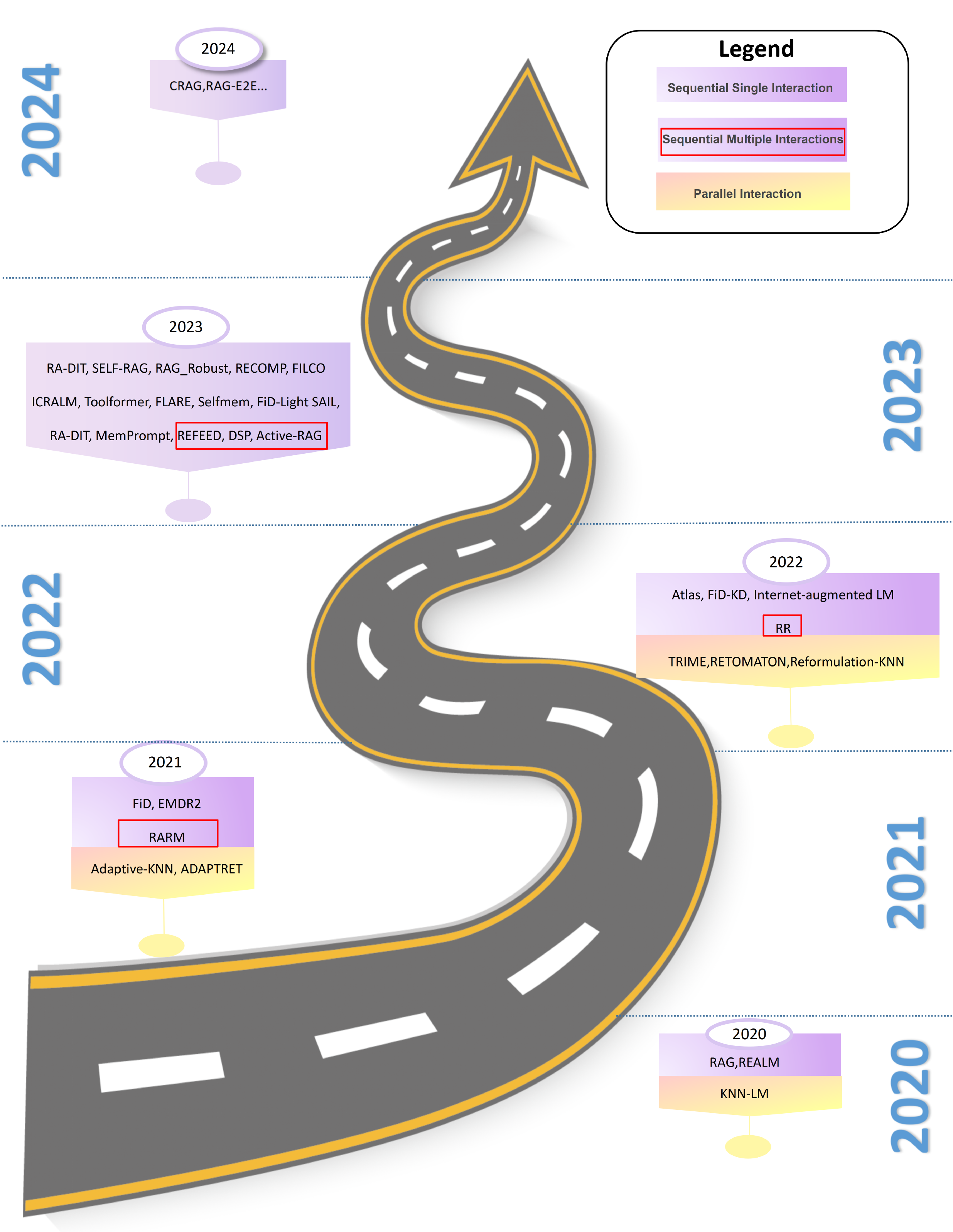}
    \caption{A roadmap of the three types of interactions. The purple areas represent work on Sequential Interaction RALM models, the red boxes signify work on Sequential Multiple Interactions RALMs models, and the yellow areas indicate work on Parallel Interaction RALM models.}
    \vspace{-10pt}
    \label{defination2}
\end{figure*}

Retrieval-Augmented Language Model (RALM) is the process of refining the output of the LM with retrieved information to obtain a satisfactory result for the user. This section provides a detailed definition of the different modes of RALM by categorising the ways in which the retriever interacts with the language model. The specific categorization of interactions can be seen in Figure \ref{defination}. In addition, the development history of each interaction method can be seen in Figure \ref{defination2}. Assuming that $z$ is the retrieved message, $x$ is the input, $y$ is the output, and $F()$ is a function, either a language model or a data processing function, with $x$ and $z$ as independent variables, the basic architecture of RALM is defined as follows:
\begin{equation}
    y = F(x, z)
\end{equation}

\subsection{Sequential Single Interaction}
The sequential single interaction process involves finding the Top-K relevant documents $z$ to input $x$ through a retriever $P_{\eta }(z|x)$, where $\eta$ is a parameter of the retriever. Subsequently, the language model $P_{\theta  }(y_{i}|x,z,y_{r})$ receives input $x$ along with relevant documents $z$ and outputs the i-th token $y_{i}$. Parameter $\theta$ is used, along with relevant output tokens $y_{r}$. The number of relevant output tokens is related to the location and type of language model. The RALM for sequential single interaction is defined as follows:
\begin{equation}
    y_{i} = LM(z, x, y_{r})
\end{equation}
When RALM was first proposed, many researchers used this method because it aligned with their original ideas, particularly those of \citet{19}, \citet{11}, and \citet{20}.

\subsection{Sequential Multiple Interactions}
As RALM technology develops, researchers have discovered that a single interaction is insufficient for long dialogue generation and solving multi-hop problems. Therefore, a method with multiple interactions between a retriever and a language model has been proposed. In this method, the researcher includes step s and typically has the language model generate the output first. When a retrieval technique is necessary, the outputted content is used for retrieval and the relevant formulas are expressed as follows:

\begin{equation}
    y_{s} = LM(z, x|y_{<s})
\end{equation}
where $y_{s}$ represents the generated tokens at the current step $s$. Among the researchers who have employed this method, the most renowned are \citet{2}, \citet{40}, and \citet{42}.

\subsection{Parallel Interaction}
In all of the previously mentioned approaches, the flow of information has a clear sequential structure, whether from the retriever to the language model or from the language model to the retriever. However, this sequential structure may not be optimal in all domains and may be less extensible, it is important to consider alternative approaches.
Researchers have proposed a novel parallel structure in which the retriever and the language model work independently for the user input $x$. The output $y$ is then determined by weighted interpolation. $I()$ is the interpolation function. The relevant equations are expressed as follows:
\begin{equation}
    y = I(LM(x, y_{r}), z)
\end{equation} 
the specific interpolation function is:
\begin{equation}
    p(y|x) =\lambda p_{R}(y|x) + (1-\lambda ) p_{LM}(y|x)
\end{equation}
where the retrieved output tokens are denoted by $p_{R}(y|x)$, the language model generated output tokens are denoted by $p_{LM}(y|x)$, and the weights are denoted by $\lambda$. Among the researchers who have employed this method, the most renowned are \citet{22}, \citet{16}, and \citet{30}.

\begin{table*}[ht]
\begin{center}
\vspace{-10pt}
\caption{Summary of Retrievers in RALM works.}
\vspace{-10pt}
\label{Retrievers}
\renewcommand{\arraystretch}{1.11}
\scalebox{0.63}[0.63]{
\begin{tabular*}{1.4\linewidth}{|cc|l|c|l|}
\hline
\multicolumn{2}{|c|}{Category}                                                                             & Technique                             & Year & Reference                        \\ \hline
\multicolumn{1}{|c|}{}                                    &                                                &                                       & 2023 & \cite{2}                         \\ \cline{4-5} 
\multicolumn{1}{|c|}{}                                    &                                                &                                       & 2023 & \cite{13}                        \\ \cline{4-5} 
\multicolumn{1}{|c|}{}                                    &                                                &                                       & 2022 & \cite{15}                        \\ \cline{4-5} 
\multicolumn{1}{|c|}{}                                    &                                                &                                       & 2024 & \cite{17}                        \\ \cline{4-5} 
\multicolumn{1}{|c|}{}                                    &                                                &                                       & 2020 & {\color[HTML]{882D00} \cite{20}} \\ \cline{4-5} 
\multicolumn{1}{|c|}{}                                    &                                                &                                       & 2022 & \cite{23}                        \\ \cline{4-5} 
\multicolumn{1}{|c|}{}                                    &                                                &                                       & 2024 & \cite{25}                        \\ \cline{4-5} 
\multicolumn{1}{|c|}{}                                    &                                                &                                       & 2023 & \cite{29}                        \\ \cline{4-5} 
\multicolumn{1}{|c|}{}                                    &                                                &                                       & 2022 & \cite{40}                        \\ \cline{4-5} 
\multicolumn{1}{|c|}{}                                    & \multirow{-10}{*}{Word Frequency}              & \multirow{-10}{*}{BM25\cite{52}}      & 2023 & \cite{44}                        \\ \cline{2-5} 
\multicolumn{1}{|c|}{}                                    &                                                &                                       & 2022 & \cite{16}                        \\ \cline{4-5} 
\multicolumn{1}{|c|}{}                                    &                                                &                                       & 2022 & \cite{21}                        \\ \cline{4-5} 
\multicolumn{1}{|c|}{}                                    &                                                & \multirow{-3}{*}{KNN search}          & 2019 & \cite{22}                        \\ \cline{3-5} 
\multicolumn{1}{|c|}{}                                    &                                                & GUD-IR\cite{38}                       & 2022 & \cite{38}                        \\ \cline{3-5} 
\multicolumn{1}{|c|}{}                                    &                                                & GAR\cite{132}                         & 2020 & \cite{132}                       \\ \cline{3-5} 
\multicolumn{1}{|c|}{\multirow{-16}{*}{Sparse Retrieval}} & \multirow{-6}{*}{Sparse Vector Representation} & Spider\cite{61}                       & 2023 & \cite{13}                        \\ \hline
\multicolumn{1}{|c|}{}                                    &                                                &                                       & 2022 & \cite{5}                         \\ \cline{4-5} 
\multicolumn{1}{|c|}{}                                    &                                                &                                       & 2021 & \cite{6}                         \\ \cline{4-5} 
\multicolumn{1}{|c|}{}                                    &                                                & \multirow{-3}{*}{COLBERTV2\cite{55}}  & 2023 & \cite{8}                         \\ \cline{3-5} 
\multicolumn{1}{|c|}{}                                    &                                                &                                       & 2023 & \cite{9}                         \\ \cline{4-5} 
\multicolumn{1}{|c|}{}                                    &                                                &                                       & 2021 & \cite{10}                        \\ \cline{4-5} 
\multicolumn{1}{|c|}{}                                    &                                                & \multirow{-3}{*}{Contriever\cite{53}} & 2020 & \cite{11}                        \\ \cline{3-5} 
\multicolumn{1}{|c|}{}                                    &                                                & DBE\cite{37}                          & 2023 & \cite{13}                        \\ \cline{3-5} 
\multicolumn{1}{|c|}{}                                    &                                                & DKR\cite{6}                           & 2023 & \cite{14}                        \\ \cline{3-5} 
\multicolumn{1}{|c|}{}                                    &                                                &                                       & 2020 & \cite{19}                        \\ \cline{4-5} 
\multicolumn{1}{|c|}{}                                    &                                                &                                       & 2020 & \cite{20}                        \\ \cline{4-5} 
\multicolumn{1}{|c|}{}                                    &                                                &                                       & 2020 & \cite{34}                        \\ \cline{4-5} 
\multicolumn{1}{|c|}{}                                    &                                                &                                       & 2022 & \cite{23}                        \\ \cline{4-5} 
\multicolumn{1}{|c|}{}                                    &                                                & \multirow{-5}{*}{DPR\cite{34}}        & 2020 & \cite{37}                        \\ \cline{3-5} 
\multicolumn{1}{|c|}{}                                    &                                                & GTR\cite{54}                          & 2019 & \cite{35}                        \\ \cline{3-5} 
\multicolumn{1}{|c|}{}                                    &                                                & E2E-NR\cite{10}                     & 2021 & \cite{30}                        \\ \cline{3-5} 
\multicolumn{1}{|c|}{}                                    &                                                & REALM\cite{11}                     & 2023 & \cite{28}                        \\ \cline{3-5} 
\multicolumn{1}{|c|}{}                                    &                                                & ORQA\cite{35}                     & 2022 & \cite{42}                        \\ \cline{3-5} 
\multicolumn{1}{|c|}{}                                    & \multirow{-18}{*}{Word Embedding}              & EMDR2\cite{36}                     & 2021 & \cite{55}                        \\ \cline{2-5} 
\multicolumn{1}{|c|}{}                                    &                                                & MuRAG\cite{18}                     & 2022 & \cite{18}                        \\ \cline{3-5} 
\multicolumn{1}{|c|}{}                                    &                                                & RA-CM3\cite{78}                     & 2022 & \cite{78}                        \\ \cline{3-5} 
\multicolumn{1}{|c|}{}                                    &                                                & RE-IMAGEN\cite{48}                     & 2022 & \cite{48}                        \\ \cline{3-5} 
\multicolumn{1}{|c|}{}                                    &                                                & MDTIG\cite{49}                     & 2022 & \cite{49}                        \\ \cline{3-5} 
\multicolumn{1}{|c|}{}                                    & \multirow{-5}{*}{Multimodal Retrieval}         & RDM\cite{50}                     & 2022 & \cite{50}                        \\ \cline{2-5} 
\multicolumn{1}{|c|}{}                                    &                                                & DRAGON\cite{56}                       & 2023 & \cite{26}                        \\ \cline{3-5} 
\multicolumn{1}{|c|}{\multirow{-25}{*}{Dense Retrieval}}  & \multirow{-2}{*}{Knowledge Distillation}            & REPULG LSR\cite{41}                   & 2023 & \cite{41}                        \\ \hline
\multicolumn{2}{|c|}{}                                                                                     & FLARE\cite{2}                      & 2023 & \cite{2}                         \\ \cline{3-5} 
\multicolumn{2}{|c|}{}                                                                                     & RAG-Robust\cite{28}                     & 2023 & \cite{28}                        \\ \cline{3-5} 
\multicolumn{2}{|c|}{}                                                                                     & IADG\cite{57}                     & 2021 & \cite{57}                        \\ \cline{3-5} 
\multicolumn{2}{|c|}{\multirow{-4}{*}{Internet Retrieval}}                                                 & Webgpt\cite{58}                     & 2021 & \cite{58}                        \\ \hline
\multicolumn{2}{|c|}{}                                                                                     & DuckDuckGo+BM25\cite{27}              & 2023 & \cite{27}                        \\ \cline{3-5} 
\multicolumn{2}{|c|}{}                                                                                     & Internet+TF-IDF\cite{39}              & 2022 & \cite{39}                        \\ \cline{3-5} 
\multicolumn{2}{|c|}{}                                                                                     & REVEAL \cite{43}                     & 2023 & \cite{43}                        \\ \cline{3-5} 
\multicolumn{2}{|c|}{}                                                                                     & NAG-ERE\cite{45}                     & 2018 & \cite{45}                        \\ \cline{3-5} 
\multicolumn{2}{|c|}{}                                                                                     & Internet+BM25\cite{59}                & 2021 & \cite{59}                        \\ \cline{3-5} 
\multicolumn{2}{|c|}{\multirow{-6}{*}{Hybrid Retrieval}}                                                    & kNN+BM25+translation model\cite{60}   & 2016 & \cite{60}                        \\ \hline
\end{tabular*}
}
\vspace{-10pt}
\end{center}
\end{table*}

\section{Retriever}
\label{s3}
Retrievers play a crucial role in the RALM architecture. The information obtained through retrievers can significantly improve the accuracy of the LM. This section provides a summary of the retrieval methods commonly used in the RALM architecture. The retrieval methods are classified into four categories based on their methods and sources: Sparse Retrieval, Dense Retrieval, Internet Retrieval, and Hybrid Retrieval. Table \ref{Retrievers} lists information about specific applications of retrievers in the RALM.

\subsection{Sparse Retriever}
For a period of time following the proposal of the retrieval technique, sparse retrieval proves to be a straightforward and effective tool in solving problems, particularly those based on knowledge. One of the main advantages of sparse retrieval is its simplicity, which can be easily integrated into existing indexing systems due to the fewer dimensions involved. \cite{51}This is consistent with human cognitive processes. Additionally, sparse retrieval is easier to generalise and more efficient. Sparse retrieval used in RALM can be classified into two categories: Word Frequency and Sparse Vector Representation. The choice between the two depends on whether machine learning is used.

\subsubsection{Word Frequency}
In the initial stage, individuals often use methods for retrieval that involve matching of relevant content, such as the TF-IDF \cite{ramos2003using} and BM25 \cite{52} algorithms, which are considered classic and effective. 

The TF-IDF algorithm utilises term frequency (TF) and inverse document frequency (IDF) to represent relevance, which has the advantages of simplicity and speed, and even if the corpus is unchanged, the TF-IDF value for each word can be computed in advance. In RALM, \citet{39} utilise the TF-IDF algorithm to match information obtained from user queries and calls to the Google search API. \citet{45} also employ the algorithm to score the generated results. The BM25 represents an enhancement over the TF-IDF. It considers the user's query and calculates the relevance score as the weighted sum of the relevance of each query word to the document. The IDF algorithm is used to derive the weight of each word, but it is improved by two moderating factors to prevent the strength of the influence of a certain factor from being infinite. This is consistent with common sense. Due to its excellent generalisation capabilities, many Retrieval-Augmented Language Model (RaLM) architectures, particularly those oriented towards open domains, employ BM25 as a retrieval method, such as \citet{2}, \citet{13} and \citet{15}. 

\subsubsection{Sparse Vector Representation}
It has become evident that simple term matching is no longer sufficient to meet the demand. Manual labelling can solve problems such as the synonym issue, but it is a resource-intensive method. With the rise of machine learning, sparse vectors are now used to represent words and retrieve them by calculating the distance between them. \cite{51} Sparse vector representation techniques differ from term matching methods in that they construct sparse vectors for queries and documents. The purpose of these representations is to capture the semantic essence of each input text, which places queries and documents in a latent space. 

\citet{61} utilised the fact that when given two paragraphs with the same repeat span, one was used to construct a query and the other as the retrieval target. The remaining paragraph in the document, which did not contain a repeat span, was used as a negative example and \citet{13} applied this retriever for the first time to the RALM architecture. On the other hand, both \citet{132} and \citet{38} proposed using language models (LM) to enhance retrieval accuracy of sparse vector representation by generating new queries from a given question and using them to retrieve relevant documents. However, Mao's approach emphasizes query expansion, while Maddan's approach emphasizes understanding the user's input.  

\subsection{Dense Retriever}
The emergence of deep learning techniques has significantly transformed the field of retrieval. There is a growing interest in using deep learning techniques to enhance retrieval accuracy, even if it means sacrificing some level of comprehensibility. The dual encoder architecture is a common design for dense retrieval models. \cite{51} The system comprises of two distinct networks that receive separate inputs, namely queries and documents, and independently generate dense embeddings for each input. Due to its high accuracy and dual-encoder structure, which is more suitable for RALM, most articles choose to use the dense indexing method to build their retrievers. This section classifies dense retrieval into three types: Word Embedding, Multimodal Retrieval, and Data Distillation, based on the characteristics of each retrieval method. 

\subsubsection{Word Embedding}
Word embeddings are a common approach in natural language processing. Similar to sparse vector representations, they use deep learning techniques to project words into a higher-dimensional vector space. Several articles in the RALM architecture utilize this technique, and we have selected representative ones to describe. 

\citet{34} proposed the DPR retrieval model, which indexes all passages in a low-dimensional and continuous space. This allows the reader to efficiently retrieve the first k passages associated with the input problem at runtime. A dense encoder is used to map any text passage to a d-dimensional real-valued vector, creating an index for all M passages used for retrieval. Due to its excellent performance as a retriever in RALM architectures, DPR has been widely adopted by researchers such as \citet{19}, \citet{20}, and \citet{34}. \citet{37} takes a similar tactic, unlike DPR, in that he uses the same encoding function for questions and paragraphs through shared parameters. In order to further minimise the intervention and reduce the cost of manual annotation, \citet{53} proposed another retriever called Contriever, which was trained using unsupervised data. It is based on successive dense embeddings and has a dual-encoder architecture. Average pooling was applied on the output of the previous layer to obtain one vector representation for each query or document. The similarity score between the query and each document was obtained by computing the dot product between their corresponding embeddings. Researchers \cite{9}, \cite{10}, and \cite{11} have used it as a retriever in the RALM architecture due to its ability to utilize unsupervised data.

\subsubsection{Multimodal Retrieval}
Retrieval techniques for multimodal tasks are more complex than those for text-only tasks \cite{80}. This is because they involve the inter-transformation of information about different states. For example, in the image-text domain, multimodal techniques for dense text retrieval have attracted interest as a means of bridging the gap between different modalities \cite{zhao2024concentrated,zhao2024dual}. Researchers have developed methods to encode textual and visual information into a shared latent space for retrieval tasks. 

\citet{49} designed four matching algorithms for handling multimodal tasks, namely sentence-to-sentence, sentence-to-image, word-to-word and word-to-image. They used reweighting for more accurate correlation calculations and cosine similarity scores as a criterion to explore the effectiveness of each algorithm. Unlike the former, \citet{78} utilised a straightforward extension of CLIP \cite{133} to divide a multimodal document into text and image components, which were encoded with frozen encoders. The L2 norm was scaled to 1 using an average pooling technique, resulting in a vector representation of the document. Similarly to \citet{49}, they also employed a Maximum Inner Product Search (MIPS) as relevance scores, ultimately selecting K documents. 

\subsubsection{Knowledge Distillation}
Knowledge distillation is a technique for gradually filtering and streamlining a large database to make it more suitable for a user's query. \cite{gou2021knowledge} involves transferring data from a pre-trained, larger model to a smaller one, often using methods such as embedded matching. Research has even been conducted on data distillation using LMs as the technology has evolved. 

\citet{41} utilises knowledge distillation to divide the retrieval process into four distinct steps. Firstly, documents are retrieved and retrieval likelihoods are computed. Secondly, the retrieved documents are scored using a language model. Thirdly, the parameters of the retrieval model are updated by minimising the KL discrepancy between the retrieval likelihoods and the distribution of the language model scores. Finally, asynchronously updating of the indexes of the data is performed. Based on this technique, \citet{56} further improve the accuracy of knowledge distillation. They present a data distillation scheme that combines sentence truncation and query enhancement with incremental relevance label enhancement using multiple enhancers.

\subsection{Internet Retrieval}
With the advancement of Internet search and sorting technology, some researchers have focused their search efforts on Internet retrieval, which is a plug-and-play approach. This approach allows non-specialists to benefit from RALM and is better suited to the open domain and generalisation. Another advantage of this retrieval model is that it does not require real-time updating of the database, but relies on updates from commercial search engines. However, despite the advantages of simplicity and convenience, there is a significant amount of irrelevant and even harmful information on the Internet that can hinder the work of RALM. If an effective screening mechanism is not implemented, the effectiveness of RALM will be significantly reduced. 

Unlike most studies \cite{28} \cite{58} that directly utilise commercial search engine APIs. \citet{57} propose an alternative approach to using multiple commercial search engine APIs. They suggest using the Bing Search API to generate a list of URLs for each query. These URLs are then used as keys to look up their page content in a lookup table constructed from public crawl snapshots, which populates a set of pages for that query. In addition, the evaluation takes into account whether the URL is from the English Wikipedia. If so, the page title is extracted from the URL and the corresponding page is searched for in the Wikipedia dump.

\subsection{Hybrid Retrieval}
As researchers gain a better understanding of the strengths and weaknesses of various retrieval techniques, they are increasingly opting to combine them, as described above. This is done in the hope of further exploiting the advantages of these techniques to improve the effectiveness and robustness of the RALM architecture.

To tackle the issue of inaccurate Internet retrieval results, \citet{39} proposed using the TF-IDF algorithm to score the retrieval results. They used each question q verbatim as a query and issued a call to Google Search via the Google Search API. For each question, they retrieved the top 20 URLs and parsed their HTML content to extract clean text, generating a set of documents D for each question q. To prevent irrelevant information from hindering the resolution of a user's query, \citet{43} designed a gating circuit. This circuit utilised a dual-encoder dot product to calculate similarity and a gating circuit based on term weights. Additionally, \citet{60} presented an approach that replaced term-based retrieval with k-Nearest Neighbors(kNN) search while combining a translation model and BM25 to improve retrieval performance. This approach enabled the model to take into account the semantic relationships between terms and traditional statistical weighting schemes, resulting in a more efficient retrieval system.

\begin{table*}[ht]
\begin{center}
\vspace{-10pt}
\caption{Summary of LMs in RALM methods.}
\vspace{-10pt}
\label{LM}
\renewcommand{\arraystretch}{1.1}
\scalebox{0.74}[0.74]{
\begin{tabular*}{1.22\linewidth}{|cc|l|c|l|}
\hline
\multicolumn{2}{|c|}{Category}                                                                        & Technique                          & \multicolumn{1}{l|}{Year} & Reference \\ \hline
\multicolumn{2}{|c|}{\multirow{3}{*}{AutoEncoder Language Model}}                                     & \multirow{2}{*}{RoBERTa\cite{110}} & 2021                      & \cite{6}  \\ \cline{4-5} 
\multicolumn{2}{|c|}{}                                                                                &                                    & 2024                      & \cite{17} \\ \cline{3-5} 
\multicolumn{2}{|c|}{}                                                                                & BERT\cite{90}                      & 2021                      & \cite{10} \\ \hline
\multicolumn{1}{|c|}{\multirow{24}{*}{AutoRegressive Language Model}} & \multirow{12}{*}{GPT Family}  & \multirow{2}{*}{GPT-3.5}           & 2023                      & \cite{2}  \\ \cline{4-5} 
\multicolumn{1}{|c|}{}                                                &                               &                                    & 2022                      & \cite{42} \\ \cline{3-5} 
\multicolumn{1}{|c|}{}                                                &                               & GPT-2\cite{99}                     & 2023                      & \cite{13} \\ \cline{3-5} 
\multicolumn{1}{|c|}{}                                                &                               & GPT-Neo\cite{91}                   & 2022                      & \cite{23} \\ \cline{3-5} 
\multicolumn{1}{|c|}{}                                                &                               & chatGPT                            & 2023                      & \cite{24} \\ \cline{3-5} 
\multicolumn{1}{|c|}{}                                                &                               & \multirow{2}{*}{GPT-4\cite{88}}    & 2023                      & \cite{4}  \\ \cline{4-5} 
\multicolumn{1}{|c|}{}                                                &                               &                                    & 2023                      & \cite{27} \\ \cline{3-5} 
\multicolumn{1}{|c|}{}                                                &                               & \multirow{2}{*}{GPT-3\cite{104}}   & 2022                      & \cite{38} \\ \cline{4-5} 
\multicolumn{1}{|c|}{}                                                &                               &                                    & 2022                      & \cite{40} \\ \cline{3-5} 
\multicolumn{1}{|c|}{}                                                &                               & GPT\cite{134}                      & 2023                      & \cite{41} \\ \cline{3-5} 
\multicolumn{1}{|c|}{}                                                &                               & \multirow{2}{*}{GPT-J}             & 2024                      & \cite{25} \\ \cline{4-5} 
\multicolumn{1}{|c|}{}                                                &                               &                                    & 2024                      & \cite{46} \\ \cline{2-5} 
\multicolumn{1}{|c|}{}                                                & \multirow{6}{*}{Llama Family} & \multirow{4}{*}{Llama2\cite{86}}   & 2024                      & \cite{3}  \\ \cline{4-5} 
\multicolumn{1}{|c|}{}                                                &                               &                                    & 2023                      & \cite{4}  \\ \cline{4-5} 
\multicolumn{1}{|c|}{}                                                &                               &                                    & 2023                      & \cite{14} \\ \cline{4-5} 
\multicolumn{1}{|c|}{}                                                &                               &                                    & 2023                      & \cite{28} \\ \cline{3-5} 
\multicolumn{1}{|c|}{}                                                &                               & \multirow{2}{*}{Llama\cite{95}}    & 2023                      & \cite{26} \\ \cline{4-5} 
\multicolumn{1}{|c|}{}                                                &                               &                                    & 2023                      & \cite{27} \\ \cline{2-5} 
\multicolumn{1}{|c|}{}                                                & \multirow{6}{*}{Others}       & Alpaca\cite{87}                    & 2024                      & \cite{3}  \\ \cline{3-5} 
\multicolumn{1}{|c|}{}                                                &                               & \multirow{2}{*}{OPT\cite{93}}      & 2023                      & \cite{13} \\ \cline{4-5} 
\multicolumn{1}{|c|}{}                                                &                               &                                    & 2023                      & \cite{26} \\ \cline{3-5} 
\multicolumn{1}{|c|}{}                                                &                               & XGLM\cite{96}                      & 2024                      & \cite{17} \\ \cline{3-5} 
\multicolumn{1}{|c|}{}                                                &                               & BLOOM\cite{106}                    & 2023                      & \cite{41} \\ \cline{3-5} 
\multicolumn{1}{|c|}{}                                                &                               & Mistral\cite{103}                  & 2024                      & \cite{46} \\ \hline
\multicolumn{2}{|c|}{\multirow{16}{*}{Encoder-Decoder Language Model}}                                & \multirow{9}{*}{T5\cite{89}}       & 2022                      & \cite{5}  \\ \cline{4-5} 
\multicolumn{2}{|c|}{}                                                                                &                                    & 2023                      & \cite{8}  \\ \cline{4-5} 
\multicolumn{2}{|c|}{}                                                                                &                                    & 2021                      & \cite{10} \\ \cline{4-5} 
\multicolumn{2}{|c|}{}                                                                                &                                    & 2023                      & \cite{14} \\ \cline{4-5} 
\multicolumn{2}{|c|}{}                                                                                &                                    & 2022                      & \cite{18} \\ \cline{4-5} 
\multicolumn{2}{|c|}{}                                                                                &                                    & 2021                      & \cite{36} \\ \cline{4-5} 
\multicolumn{2}{|c|}{}                                                                                &                                    & 2020                      & \cite{37} \\ \cline{4-5} 
\multicolumn{2}{|c|}{}                                                                                &                                    & 2022                      & \cite{39} \\ \cline{4-5} 
\multicolumn{2}{|c|}{}                                                                                &                                    & 2023                      & \cite{43} \\ \cline{3-5} 
\multicolumn{2}{|c|}{}                                                                                & \multirow{7}{*}{BART\cite{12}}     & 2021                      & \cite{6}  \\ \cline{4-5} 
\multicolumn{2}{|c|}{}                                                                                &                                    & 2023                      & \cite{9}  \\ \cline{4-5} 
\multicolumn{2}{|c|}{}                                                                                &                                    & 2019                      & \cite{12} \\ \cline{4-5} 
\multicolumn{2}{|c|}{}                                                                                &                                    & 2020                      & \cite{19} \\ \cline{4-5} 
\multicolumn{2}{|c|}{}                                                                                &                                    & 2023                      & \cite{28} \\ \cline{4-5} 
\multicolumn{2}{|c|}{}                                                                                &                                    & 2020                      & \cite{37} \\ \cline{4-5} 
\multicolumn{2}{|c|}{}                                                                                &                                    & 2022                      & \cite{39} \\ \hline
\end{tabular*}}
\vspace{-10pt}
\end{center}
\end{table*}

\section{Language Models}
\label{s4}
Although humans used to rely solely on searching for information, the development of language models has revolutionised the field of natural language processing, making it more vibrant and creative. In contrast to LM, which employs solely the parameters derived from training to complete the task, RALM integrates the nonparametric memory acquired by the retriever with the parametric memory of LM itself to create a semiparametric memory, thereby enhancing the performance of the language model. In the RALM architecture, many researchers utilise off-the-shelf language models for evaluation. This section introduces the language models commonly used in RALM architectures and classifies them into three categories: AutoEncoderlanguage model, AutoRegressive language model and Encoder-Decoder model. Table \ref{LM} lists information about specific applications of LM in the RALM.

\subsection{AutoEncoder Language Model}
The logical process of an AutoEncoder is that the original input (set to x) is weighted and mapped to y, which is then inversely weighted and mapped back to z. If through iterative training the loss function L(H) is minimised, i.e. z is as close to x as possible, i.e. x is perfectly reconstructed, then it can be said that forward weighting is a successful way of learning the key features of the input. AutoEncoder language models take their name from the Denoising AutoEncoder (DAE) \cite{vincent2008extracting}, which is used to predict tokens that are [masked] by contextual words (these [masked] words are actually noise added at the input, typical of thinking). DAE is a technique that involves adding random noise to the input layer of data. This helps to learn more robust features when using an unsupervised approach to pre-train the weights of a deep network in a hierarchical manner.

Most of AutoEncoder language models are highly generalisable, unsupervised, and do not require data annotation. They can naturally incorporate contextual semantic information. However, the independence assumption introduced in the Pre-Training stage means that the correlation between predicted [MASK] is not considered. Additionally, the introduction of [Mask] as a special marker in the input to replace the original Token creates inconsistency between the data in the Pre-Training stage and the Fine-Tuning stage, where [Mask] is not present. Self-encoding language models are commonly used in RALM architectures for natural language understanding (NLU) tasks.

As AutoEncoder language models excel at Natural Language Understanding (NLU) tasks, many RALM architectures\cite{6} \cite{17} \cite{10}utilise them for specific tasks, such as judgement. One of the most commonly used models is BERT and its improved versions.\citet{110} proposed the BERT model, which was inspired by closed tasks\cite{121}. RoBERTa\cite{90} is trained using dynamic masking, full sentences without NSP loss, large mini-batches, and a larger byte-level BPE to address the lack of training of Bert's model. According to \citet{136}, BERT heavily relies on global self-attention blocks, resulting in a large memory footprint and computational cost. Although all attention heads query the entire input sequence, some only need to learn local dependencies, leading to computational redundancy. To address this issue, They proposed a new span-based dynamic convolution to replace these self-attention heads and directly model local dependencies. The new convolutional head, along with other self-attentive heads, forms a hybrid attention block. Furthermore, \citet{135} was able to decrease the size of the BERT model by 40\% while maintaining 97\% of its language comprehension abilities and achieving a 60\% increase in speed by implementing knowledge distillation during the pre-training phase.

\subsection{AutoRegressive Language Model}
The primary purpose of an AutoRegressive language model is to predict the next word based on the preceding words. This is commonly known as left-to-right language modelling, where the token at the current time t is predicted based on the first t-1 tokens.

This model has the advantage of being left-to-right, which is beneficial for generative natural language processing tasks like dialog generation and machine translation. AutoRegressive language models are well-suited to this process, making this model a popular choice for NLG tasks in the field of RALM. However, The information in question can be utilized only from the preceding or following text, and not in combination with both. OpenAI has made a notable impact on the field of research pertaining to autoregressive language models. Recently, Google has also made advancements in research on the model.

The GPT family is one of the most common examples of AutoRegressive language models. It was first proposed by \citet{134}, who identified a basic architecture of unsupervised pre-training followed by fine-tuning. \citet{99} later proposed zero-shot learning based on GPT. Later, \citet{104} proposed GPT-3 using an approach similar to \citet{99}, which involved scaling up and abandoning fine-tuning. They also utilized alternately dense and locally banded sparse attentional patterns in the transformer layer, similar to the sparse transformer \cite{122}. There are also several related studies, such as GPT-NEO\cite{91} and ChatGPT, which use Reinforcement Learning from Human Feedback (RLHF). RLHF has significantly enhanced the accuracy of GPT models. Although ChatGPT is not open source, many researchers \cite{2} \cite{42} still use its API for generative tasks in RALM. Recently, report on GPT-4\cite{88} have appeared, with the main focus on building a predictable and scalable deep learning stack dedicated to improving GPT-4's safety and alignment. Many researchers \cite{4} \cite{27} have recently used GPT-4 to generate prompts for RALM.

The Llama family is a well-known class of AutoRegressive language models. Llama\cite{95} was first proposed as a language model that uses only publicly available data. To improve training stability, they normalise the input of each transformer sub-layer instead of normalising the output. They use the RMSNorm normalisation function introduced by \citet{100} and replace the ReLU non-linearity with the SwiGLU activation function introduced by \citet{111} to improve performance. Furthermore, the authors replaced the absolute position embedding with the rotational position embedding (RoPE) introduced by \citet{116} at each layer of the network. Llama2\cite{86} used supervised fine-tuning, initial and iterative reward modelling and RLHF in their experiments. they also invented a new technique, Ghost Attention (GAtt), which helps to control the flow of dialogue in multiple turns. Qwen\cite{98}, based on Llama, the following adjustments were made: 1. The method of loose embedding was chosen instead of bundling the weights of input embedding and output projection to save memory cost. 2. The accuracy of the inverse frequency matrix was improved. 3. For most of the layers, bias was eliminated as per \citet{117}. However, bias was added in the QKV layer to enhance the model's extrapolation capability. The traditional layer normalization technique described in \citet{118} was replaced with RMSNorm. They chose SwiGLU as their activation function, which is a combination of Swish \citet{119} and gated linear units \cite{120}. The dimension of the feed-forward network (FFN) is also reduced. Furthermore, Mistral 7b \cite{103} utilises Grouped Query Attention (GQA) to enhance inference speed and combines it with Sliding Window Attention (SWA) to efficiently process sequences of any length with reduced inference cost. These techniques demonstrate superior performance over Llama2.The Llama model is open source and uses publicly available data, providing researchers \cite{3} \cite{4} \cite{14} with more opportunities to expand. As a result, many researchers use the Llama family as language models in the RALM architecture.

\subsection{Encoder-Decoder Language Model}
Transformer\cite{107} is an "encoder-decoder" architecture, which consists of encoders and decoders superimposed on multi-head self-attention modules. Among them, the input sequence is divided into two parts, the source sequence and the destination sequence. The former is input to the encoder and the latter is input to the decoder, and both sequences need to embed representation and add position information. The Transformer architecture enables parallel computation and the processing of entire text sequences simultaneously, resulting in a significant increase in model training and inference speed. 

\citet{89} introduces a unified framework for converting all text-based language problems into text-to-text format. The aim is to explore the potential of transfer learning techniques for natural language processing. In contrast to the original transformer, a simplified version of layer normalization is used, where activations are rescaled without additional biases. After applying layer normalization, a residual skip connection \cite{108} adds the input of each subcomponent to its output. \citet{109} is applied to the feed-forward network, the skip connections, the attentional weights, and the inputs and outputs of the entire stack. The T5 model has been widely used as a language model by many researchers, such as \citet{8}, \citet{10}, and \citet{36}. Additionally, \citet{94} proposed instruction tuning as an approach to improve model performance. The study focused on three aspects: the number of scaling tasks, the size of the scaled model, and the fine-tuning of chain of thought data. The results showed that larger model sizes and more fine-tuning tasks significantly improved model performance. Additionally, the study found that chain of thought(CoT) significantly improves inference level. \citet{14} used this approach to tune T5 and apply it to the RALM architecture. 

BART\cite{12} is an Encoder-Decoder model that allows for arbitrary noise transformations, as the input to the encoder does not need to align with the output of the decoder. In this case, the document is corrupted by replacing the text span with mask symbols. For pre-training, the researchers proposed five models: Token Masking, Token Deletion, Text Infilling, Sentence Permutation, and Document Rotation. For fine-tuning, the encoder and decoder are fed an uncorrupted document, and the representation of the final hidden state from the decoder is used. Many researchers\cite{6} \cite{9} \cite{12} have adopted BART as the language model in the RALM architecture due to its comprehensive and novel pre-training approach, which greatly enhances the model's robustness.

\begin{figure*}[t]
    \centering
    \includegraphics[width=\textwidth]{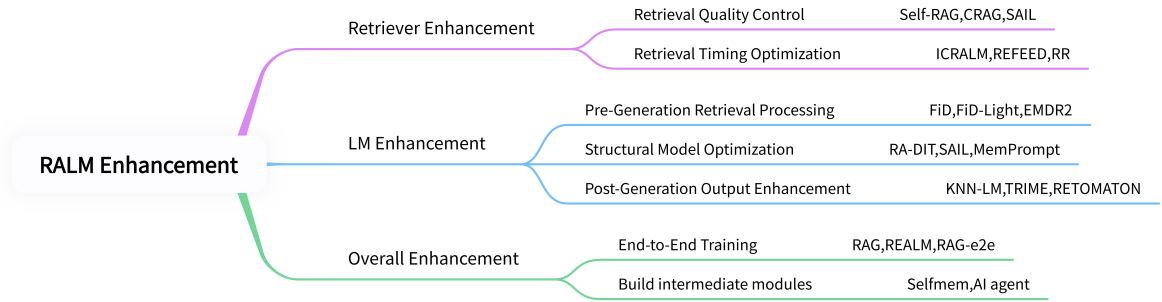}
    \caption{Classification of RALM enhancement methods.}
    \label{enhancement}
\end{figure*}

\section{RALM Enhancement}
\label{s5}
This section describes how researchers in the RALM architecture improved the output quality by enhancing its components. We divided the improvement method into three parts: Retriever Enhancement, LM Enhancement, and Overall Enhancement. Figure \ref{enhancement} illustrates the categorization of enhancement methods.
\subsection{Retriever Enhancement} 
This section presents the researchers' efforts on the retriever side, which include Retrieval Quality Control and Retrieval Timing Optimization.
\subsubsection{Retrieval Quality Control}
\citet{137} argue that retrieval can produce documents that not only fail to provide helpful information but can also compromise the quality of the language model output. As a result, many scholars in the field of RALM focus on improving the relevance between the retrieved content and the user's input to enhance the final output's quality.

\citet{26} propose an approach for instruction-tuning. They update the query encoder using a generalised LM supervised retrieval (LSR) \cite{41}training target that completes the computation through a combination of supervised tasks and unsupervised text. This enables the retriever to produce more contextually relevant results that are consistent with LLM preferences. Inspired by this instruction-tuning approach, \citet{4} proposed a more sophisticated model trained on an instruction-tracking dataset: SELF-RAG. As a result of their refinements, SELF-RAG can retrieve and select the best possible model outputs on demand through fine-grained self-reflection, making it broadly applicable, more robust, and controllable. In contrast to approaches that aim to enhance the quality of retrieved documents through the use of external models, such as natural language inference \cite{28} and summarization\cite{29} models, SELF-RAG proposes entirely novel ideas. The model divides the retrieved document into parallel segments and compares their relevance. It then combines the most similar parts of the document. \citet{3} improves on SELF-RAG by designing a correction strategy to address inaccurate retriever results. They classify the information into three categories: CORRECT, INCORRECT, and AMBIGUOUS. If the information is CORRECT, the document is refined and filtered. If it is INCORRECT, the document is discarded and the web is searched for retrieval. The term AMBIGUOUS indicates a lack of confidence in the accuracy of a judgement. In this case, a combination of the two methods mentioned above will be used. Additionally, \citet{14} proposed FILCO, a method for retrieving document content with sentence precision through three filters: STRINC, lexical overlap, and conditional cross-mutual information (CXMI).

\subsubsection{Retrieval Timing Optimization}
Researchers typically consider the timing of retrieval in two situations: when working on tasks that require multiple retrievals, such as long dialogue generation and multi-hop problems, or when it is impossible to find a suitable and relevant document. Using irrelevant documents can harm the accuracy of the output.

A simple way to determine the timing of a retrieval is to adjust the retrieval steps. \citet{13} utilised an approach that involved a prefix encoding to adjust the runtime cost. The prefix encoding of the generated content was constantly recalculated. The choice of retrieval stride is a trade-off between runtime and performance. According to Toolformer \cite{25}, the search command can be used directly to retrieve useful information when the model needs to retrieve documentation help in the process of generating content. Inspired by this idea, \citet{2} propose two methods for determining the timing of retrieval. The first method involves interrupting the generation of the LM when it encounters a place where a retrieval needs to be performed and then performing the retrieval operation. The second method involves generating a temporary sentence in its entirety. If there is a low confidence marker in the sentence, the marker is masked and the rest of the sentence is used for retrieval. \citet{44} also employed  LM to determine the timing of retrieval. However, instead of generating low-confidence markers using LM, they had LM score the output before and after retrieval.\citet{23}'s approach differed from the traditional method of having LMs generate low-confidence markers. Instead, they used Wikipedia page views as a measure of prevalence and converted knowledge triples of wiki data with varying levels of prevalence into natural language questions anchored to the original entity and relation types. This approach is more objective and avoids subjective evaluations. For tasks that required reasoning, both \citet{40} and \citet{42} used chain of thought(CoT) to determine when to perform a retrieval.

\subsection{LM Enhancement}
This section presents the researchers' efforts in language modelling, including Pre-Generation Retrieval Processing, Structural Model Optimization, and Post-Generation Output Enhancement.

\subsubsection{Pre-Generation Retrieval Processing}
The RALM architecture initially used a single document for retrieval augmentation. However, it was discovered that RALM's performance significantly improved when the number of retrieved paragraphs was increased. \cite{20} Therefore, they proposed a new method called Fusion-in-Decoder (FiD) which involves keeping the retriever unchanged, using the encoder in LM to encode the related documents one by one, and then connecting the related documents and giving them to the decoder for output. Then \citet{8} improved on the FiD. They constrained the information flow from encoder to decoder. FiD-Light with reranking was also tuned via text source pointers to improve the topmost source accuracy. \citet{37} applied knowledge distillation to the FiD model, also known as FiD-KD, using cross-attention scores from a sequence-to-sequence reader to obtain synthetic targets for the retriever. \citet{36} proposed an enhancement approach that differs from knowledge distillation in that it uses an end-to-end training approach requiring fewer documents, training cycles, and no supervised initialization compared to FiD-KD.

\subsubsection{Structural Model Optimization}
As language models continue to evolve at an accelerated pace, an increasing number of large models with high parameter counts and exceptional performance are emerging. Tuning the parameters and internal structure of these models has become increasingly difficult and inefficient, making instruction tuning more important than ever.

FLAN\cite{94} is one of the most systematic and comprehensive approaches among the many studies on instruction tuning. This approach fine-tunes the language model on the instruction-optimised dataset, scales the number of tasks and model size, and incorporates chain-of-thought data in the fine-tuning. Although the authors did not consider a specific approach to tuning instructions in RALM architecture, their work provides a valuable reference for future research. In the instruction fine-tuning of RALM, \citet{26} integrated in-context retrieval augmentation. This greatly reduces the likelihood of the language model being misled by irrelevant retrieval content. SAIL\cite{27} builds language generation and instruction tracking capabilities on complex search results generated by internal and external search engines. Using a corpus of instruction tuning, they collect search results for each training case from different search APIs and domains, and construct a search-based training set containing a triplet of (instruction, grounding information, response). In contrast to training on instruction-tuned datasets, \citet{38} and \citet{39} propose to prompt large models directly from retrieved knowledge. \citet{38} used GPT-3 to clarify memory pairings of recorded cases where the model misinterpreted the user's intention, as well as user feedback. This ensures that their system can generate enhanced prompts for each new query based on user feedback. In contrast, \citet{39} uses few-shot prompts and answer reordering to improve inference computation.

\subsubsection{Post-Generation Output Enhancement}
As defined in Section 2 on Parallel Interaction, this interaction is inspired by the K-Nearest Neighbor (KNN) LM \cite{22}. It is a paradigmatic instance of RALM, wherein the LM is employed solely to enhance the outcomes. Since the proposal of KNN-LM, many researchers have worked to optimize the model. In this section, we will describe the landmark work in detail.

The KNN-LM approach involves linearly interpolating the extended neural language model with the K-Nearest Neighbours in the pre-trained LM embedding space. \citet{15} proposed different processing for three types of memories (local, long-term and external) and added training for in-batch tokens to KNN-LM. The proposed changes aim to improve the performance of the model. Unlike KNN-LM, which only uses memory units during training, TRIME \cite{15} uses memory units during both testing and training. \citet{30} suggested that not all generated tokens need to be retrieved. Instead, a lightweight neural network can be trained to aid the KNN-LM in adaptive retrieval. Additionally, efficiency can be improved through database streamlining and dimension reduction. \citet{16} proposed RETOMATON, an unsupervised, weighted finite automaton built on top of the data store. RETOMATON is based on saving pointers between successive data store entries and clustering techniques. RETOMATON is more effective than ADAPTRET\cite{30} in improving accuracy by utilizing remaining pointers during KNN retrieval. Even without KNN retrieval, interpolation operations can still be performed using the stored previous information in the pointers, unlike ADAPTRET which solely relies on the language model. Furthermore, RETOMATON is unsupervised, requiring no additional data for training, making it more data-efficient. \citet{32} proposed using continuous cache to improve the performance of KNN-LM. This involves storing past hidden activations and accessing them at the appropriate time by dot product with present hidden activation. \citet{33} utilise an extended short-term context by caching local hidden states and global long-term memory by retrieving a set of nearest-neighbour tokens at each time step. They also design a gating function to adaptively combine multiple sources of information for prediction. Compared to KNN-LM, this method uses dynamic weights and can handle cases where interpolation is not feasible, such as when the memory output is an image, video, or sound. \citet{31} proposed a method for adjusting the interpolation weights. The weights are dynamically adjusted based on the size of the region of overlap between the retrieved stored data and the assessment set, which reflects the quality of the retrieval.

\subsection{Overall Enhancement}
This section presents the researchers' efforts on the RALM architecture as a whole, including End-to-End Training and Build intermediate modules.
\subsubsection{End-to-End Training}
Researchers have begun working on a method called end-to-end training, which aims to minimise manual intervention and focus solely on data. This method utilises deep learning and is becoming increasingly popular due to the growing amount of available data. During research on RALM architectures, many researchers tend to use end-to-end training methods to achieve better results.

\citet{19} and \citet{11} were among the first researchers to apply end-to-end training to the field of RALM. However, they differed in their approach. REALM \cite{11} used masked language training in the pre-training phase and included a retriever that can be trained end-to-end. In the fine-tuning phase, only the QA task was targeted while keeping the retriever frozen. On the other hand, RAG \cite{19} used an already trained retriever, DPR, and only employed BART for partial end-to-end training. Similar to REALM, \citet{10} present an unsupervised pre-training method that involves an inverse cloze task and masked salient spans. This is followed by supervised fine-tuning using question-context pairs. In addition, they find that the use of end-to-end trained retrievers resulted in a significant improvement in performance across tasks. \citet{36} apply end-to-end training to multi-document processing, in their proposed approach, the value of the latent variable, which represents the set of relevant documents for a given question, is estimated iteratively. This estimate is then used to update the parameters of the retriever and reader. \citet{9} describe the end-to-end optimization of RAG from previous studies and introduces an auxiliary training signal to incorporate more domain-specific knowledge. This signal forces RAG-end2end to reconstruct a given sentence by accessing relevant information in an external knowledge base. This approach has greatly improved domain adaptability.

\begin{figure*}[t]
    \centering
    \includegraphics[width=\textwidth]{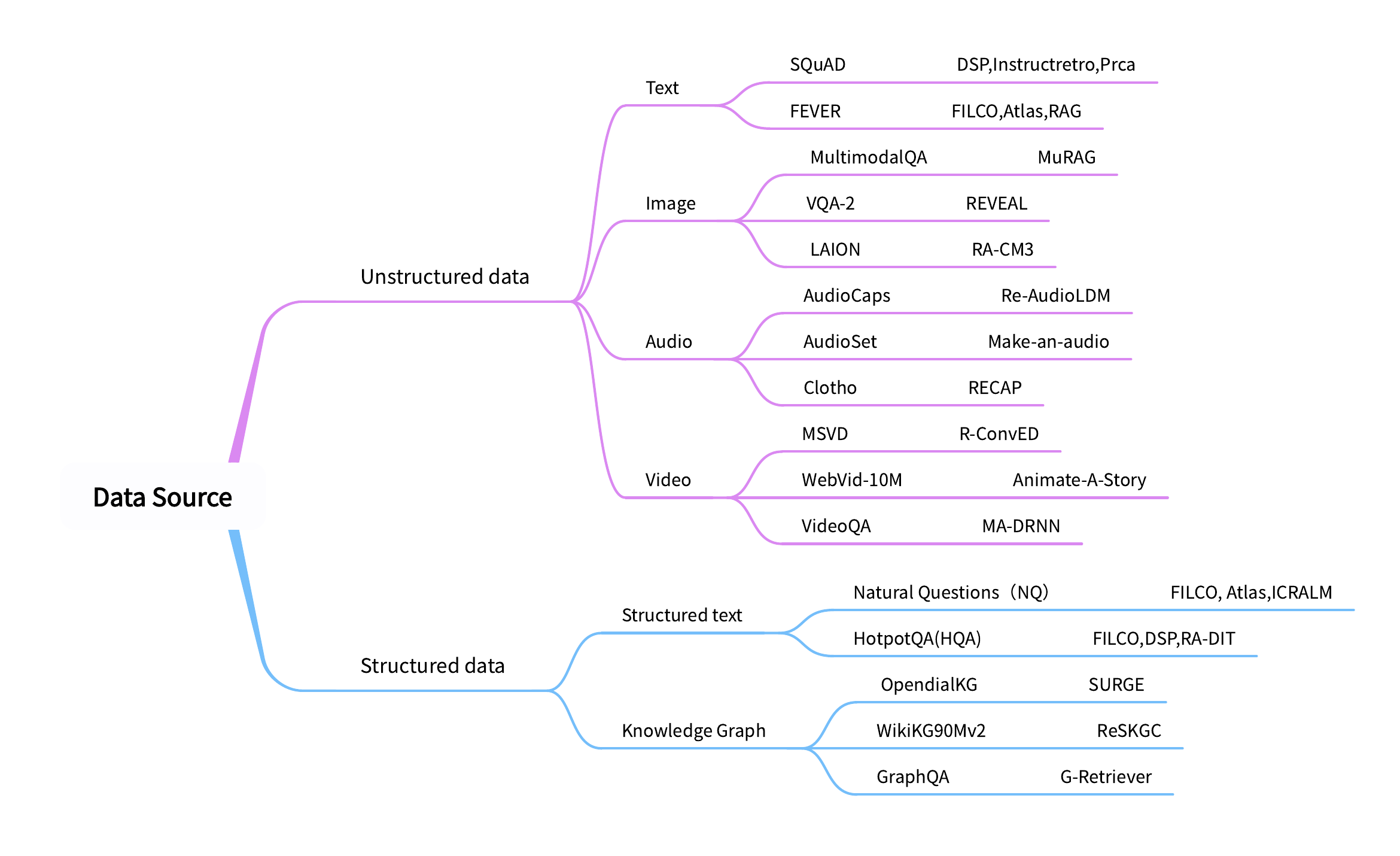}
    \caption{Classification of RALM data sources.}
    \label{data}
\end{figure*}

\subsubsection{Intermediate Modules}
Recently, some researchers have constructed an intermediate module to coordinate the activities of both the retriever and the language model due to space or black-box LLM constraints, without improving either.

\citet{17} present Selfmem, a model designed to tackle the issue of low corpus quality. Selfmem utilises a retrieval-enhanced generator to create an infinite pool of memory, which is then used by a memory selector to choose an output for subsequent generations. This approach enables the model to use its own output to enhance generation. \citet{24} propose an AI agent that formulates human system dialogue as a Markov Decision Process (MDP) described by a quintuple. The quintuple includes an infinitely large set of dialogue states, a collection of historical behaviours, a probability of state transfer, external rewards obtained, and a variable parameter.

\section{Data Sources}
\label{s6}
This section will introduce some of the data sources commonly used in RALM and categorise them into structured and unstructured data.  Figure \ref{data} illustrates the categorization of data sources.
\subsection{Structured Data}
Structured data includes various structures, such as tables and knowledge graphs. The benefit of this type of data is its clear structure, typically in tabular form, with each field precisely defined. It is appropriate for storing numbers, dates, text, and other data types. Structured data can be easily queried, analysed, and processed using a structured query language like SQL.

Natural Questions(NQ) \cite{139} is a very well-known dataset in the NLU field.The given text describes a structured question and a corresponding Wikipedia page. The page is annotated with a long answer, typically a paragraph, and a short answer consisting of one or more entities. If there is no long or short answer, it is labelled as empty. Due to the reliability of the Google search engine and its vast amount of data, many scholars have used this dataset to train RALM, such as \citet{14}, \citet{5} and \citet{13}. HotpotQA(HQA) \cite{140} stores information about multi-hop questions and provides sentence-level supporting facts needed for inference. The structure includes the paragraph, question, answer, and sentence number that supports the answer. This dataset has been used by many researchers, such as \citet{14}, \citet{42}, and \citet{147}, to train RALM for multi-hop question answering. Another significant form of structured data is the Knowledge Graph. It is a data structure that primarily consists of triples of (entities, relationships, attributes). Some of the most frequently used datasets include Wikidata5M, WikiKG90Mv2, OpendialKG, and KOMODIS. All of these models \cite{124} \cite{125} \cite{126} rely on knowledge graphs as a data source.

\subsection{Unstructured Data}
Unstructured data, in contrast, does not have a clearly defined data structure and exists in various forms, including text, images, and audio. Due to its large and diverse nature, it is challenging to store and manage in traditional tabular form. Although it contains valuable information, it requires natural language processing, image recognition, and other technologies to parse and comprehend.

Several RALM researchers, including \citet{42}, \citet{148}, and \citet{149}, have used this dataset as a source of data. The FEVER \cite{144} dataset is mainly used for fact extraction and validation. Several RALM researchers, including \citet{19}, \citet{14}, and \citet{5}, have used the factual text in this dataset as a source of data. In addition to unstructured text, there is also a significant amount of inherently less structured data, such as images, videos, and audio. Several common image datasets are available for use in research, including MNIST, CIFAR-10, Pascal VOC, and COCO. Many studies \cite{18} \cite{43} \cite{78} in the field of RALM have utilized these datasets. Common audio datasets used in speech research include LJ Speech, JSUT, and RUSLAN. Many studies \cite{151} \cite{152} \cite{153} in the field also rely on audio data as a primary source. Common video datasets used in research include HMDB, UCF101, and ASLAN. Many studies \cite{155} \cite{156} \cite{157} in the field of RALM utilize audio data as a source of information.

\begin{figure*}[t]
    \centering    
    \vspace{-20pt}
    \includegraphics[width=\textwidth]{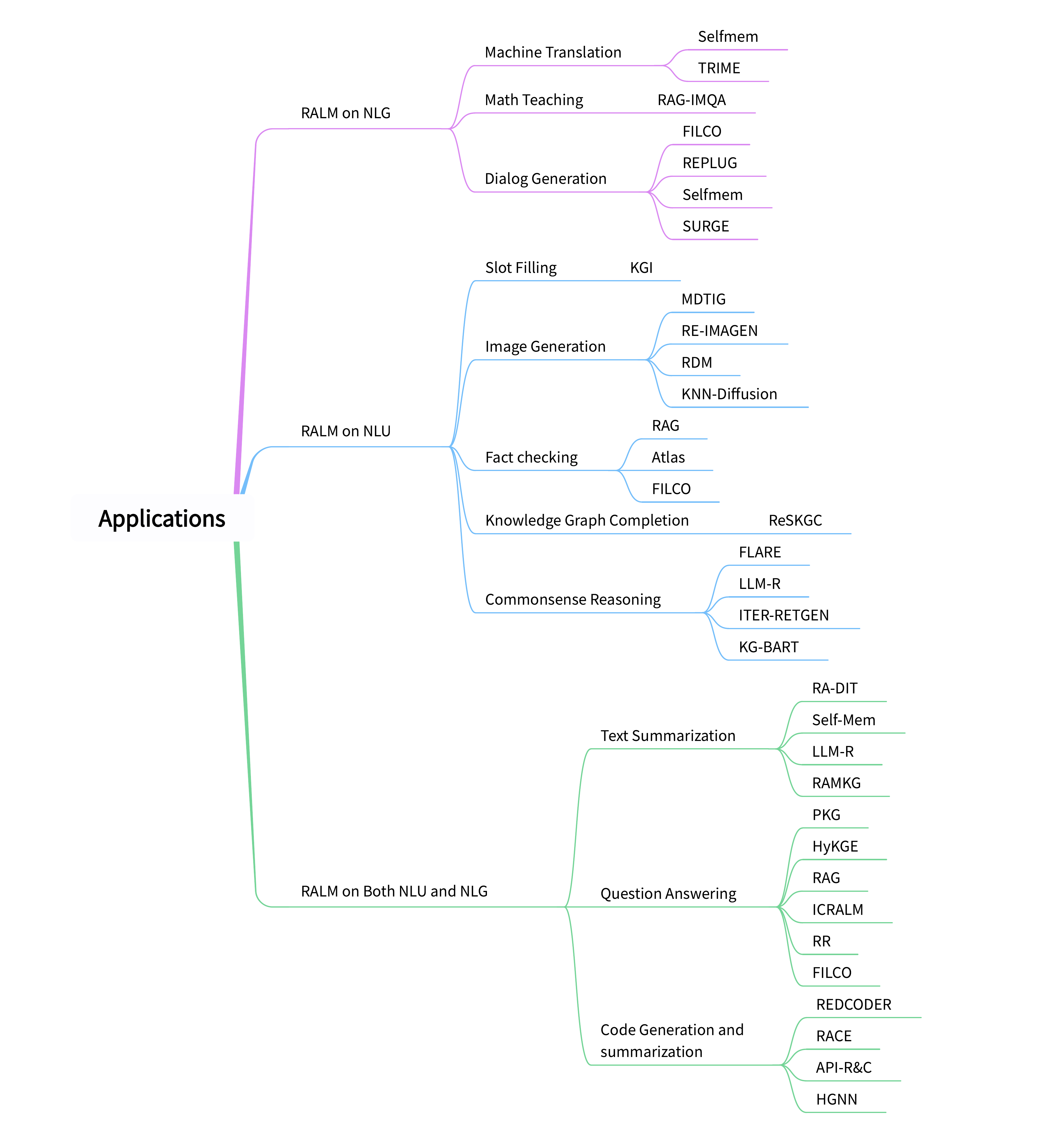}
    \vspace{-10pt}
    \caption{Classification of RALM applications.}
    \vspace{-10pt}
    \label{application}
\end{figure*}

\section{Applications}
\label{s7}
This section provides a summary of the downstream tasks that the RALM architecture primarily focuses on. The relevant application directions are categorized according to the requirements for model generation or comprehension. RALM on NLG indicates that the accomplishment of the task primarily depends on the generative capabilities. Conversely, RALM on NLU indicates that the accomplishment of the task primarily depends on the comprehension capabilities. Finally, RALM on both NLU and NLG indicates that the task is generally handled in two ways, one that relies primarily on comprehension capabilities and one that relies primarily on generative capabilities. Figure \ref{application} illustrates the categorization of applications.

\subsection{RALM on NLG Tasks}

\subsubsection{Machine Translation}
Machine translation, also known as automatic translation, is the process of converting one natural language (source language) into another natural language (target language) using a computer. It is a branch of computational linguistics, one of the ultimate goals of artificial intelligence, and has important scientific research value. Machine translation systems can be divided into two categories: rule-based and corpus-based. The former comprises a dictionary and a rule base, which collectively constitute the knowledge source. In contrast, the latter comprises a corpus that has been divided and labeled, and which does not require a dictionary or rules. Instead, it is based on statistical laws and most RALMs accomplish this task based on rules.

The Selfmem \cite{17} system employs two distinct language models for the machine translation task. The first is a trainable mini-model, which has been trained using a joint and bipartite approach, respectively. The second is a few-shot prompted LLM. Ultimately, Selfmem has demonstrated a notable enhancement in its performance across all four translation directions and for both training architectures. This outcome suggests that enhanced memory capabilities often result in superior generation outcomes. In order to achieve the best results, TRIME \cite{15} used the IWSLT'14 De-En baseline. Given that the task is sentence-level, the researchers did not use local memory and long-term memory, as there are few repetitive tokens in them. Instead, they used only external memory, which enabled them to beat the KNN-MT \cite{168} in performance.

\subsubsection{Math Teaching}
As the RALM architecture continues to evolve, an increasing number of potential application directions are being identified. Levonian \cite{69} were inspired by RALM to apply this architecture to the domain of mathematics teaching and learning. To address the fact that the knowledge stored in the LLM may not match what is taught in schools, they used one of three prompted instructional conditions to generate responses to math student queries using a retrieval enhancement generation system. Survey respondents ranked the responses according to preference and evaluated basic math textbooks as a retrieval corpus.

\subsubsection{Dialog Generation}
Dialog generation, in particular lengthy dialogue, is a challenging task. This is due to the necessity of not only ensuring that the language model possesses natural language processing capabilities, but also that the model is able to utilise context in order to satisfy the requirements of the dialogue. 

FILCO \cite{14} employs the Wikipedia dataset from the KILT benchmark, referred to as the "Wizard of Wikipedia" (WoW), to generate subsequent dialogue. This process involves basing the output on a Wikipedia article, with the input comprising the history of multiple rounds of dialogue. RA-DIT \cite{26} also employs the WoW dataset in the fine-tuning phase. As a result of the command tuning operation, the model outperforms Llama \cite{95} and Llama-REPLUG \cite{41} with the same parameters for dialogue generation in the zero-shot condition. The incorporation of Selfmem \cite{17} into the retrieval-augmented generator markedly enhances the generation of dialogue, as a consequence of its remarkable flexibility. This is achieved by the direct optimisation of memory for the desired properties of diverse and information-rich dialogues. In contrast, SURGE \cite{124} employs the Knowledge Graph as a data source for the dialogue generation task, wherein each dialogue round comprises facts from a large-scale KG. In contrast to other related work \cite{163}, they retrieve only contextually relevant subgraphs, thus avoiding the computational overheads and misleading models that can result from retrieving irrelevant data. 

\subsection{RALM on NLU Tasks}

\subsubsection{Slot Filling}
Slot filling is a technique employed in natural language processing for the purpose of recognizing and extracting specific information from user-supplied text or speech. In slot filling, the system defines a set of slots in advance, with each slot representing a specific information requirement. These requirements may include, but are not limited to, date, time, location, and so forth. Upon receipt of a user input in the form of text or speech, the system performs an analysis of the content, attempting to identify information that matches the predefined slots or classification labels \cite{lu2023medical, lu2023medkpl}. This information is then populated into the corresponding slots for subsequent processing and response.

KGI \cite{81} enhances dense channel retrieval through the utilization of hard negatives in dense indexing and implements a robust training process for retrieval enhancement generation. The retrieval enhancement is employed to enhance the effectiveness of the slot-filling task, thereby facilitating the generation of high-quality knowledge graphs by AI. The results demonstrate that the method achieves excellent performance in TREx and zsRE datasets and exhibits remarkable robustness in TACRED dataset.

\subsubsection{Image Generation}
The process of text-to-image generation is a challenging one, requiring a model to demonstrate a high degree of natural language understanding and to convey this understanding through an image, in contrast to the typical format of textual data.

In a pioneering study, \citet{49} proposed the use of retrieval techniques to enhance the quality of text-to-image generation. They conducted a comparative analysis of the quality and quantity of the generated images with mainstream models on the CUB and COCO datasets. Their findings demonstrated that all models outperformed their contemporaries. In contrast, RE-IMAGEN \cite{48} focused on assisting the model in generating images of uncommon objects through retrieval. This approach ultimately led to the achievement of exceptionally high FID scores on the COCO and WikiImage datasets. Even more groundbreaking results were obtained on the authors' own proposed EntityDrawBench benchmark, which encompasses a range of common and rare objects across multiple categories. RDM \cite{50}, although trained in a similar manner to RE-IMAGEN, employs image features as the foundation for retrieval and is supplanted by user examples during the inference process. Consequently, RDM is capable of efficiently transferring the described artistic style to the generated images. Furthermore, in contrast to RE-IMAGEN, which employs image-text pairs for retrieval, KNN-Diffusion \cite{84} solely utilizes images for retrieval, resulting in a lower quality of results on the COCO dataset. 

\subsubsection{Fact checking}
Fact checking involves verifying a statement based on evidence. This task at hand involves a retrieval problem and a challenging implicit reasoning task. Furthermore, This task typically involves taking the statement as input and producing relevant document passages that prove or disprove the statement. Many RALM models get excellent performance because they come with their own retrievers. It is an important aspect of natural language understanding. 

RAG \cite{19} uses the FEVER dataset to map labels (Supported, Refuted, NotEnoughInfo) to individual output tokens. It is trained directly using the declaration class, which is not supervised over the retrieved evidence, unlike other works. Atlas \cite{5} employs few-shot  learning to achieve performance comparable to previous studies in just 64-shot conditions. Furthermore, after training with the full dataset, it outperformed the best model \cite{162} available at the time. FILCO \cite{14} approached the task of improving the quality of retrieved documents by using the FEVER dataset from the KILT base aggregation, which only included the supports and refutes tags. Accuracy was used as a metric.

\subsubsection{Knowledge Graph Completion}
A multitude of previous tasks have employed structured data in the form of knowledge graphs, with knowledge graph completion representing a pervasive application. The conventional methodology \cite{169} \cite{170} for completion entails defining the task as a sequence-to-sequence process, wherein incomplete triples and entities are transformed into text sequences. However, this approach is constrained by its reliance on implicit reasoning, which significantly constrains the utility of the knowledge graph itself. ReSKGC \cite{125} proposes the integration of retrieval augmentation techniques with knowledge graph completion. This integration entails the selection of semantically relevant triples from the knowledge graph and their utilization as evidence to inform the generation of output through explicit reasoning. This model employs data from the Wikidata5M and WikiKG90Mv2 datasets, demonstrating superior performance compared to other existing work in a range of conditions.

\subsubsection{Commonsense Reasoning}
Commonsense Reasoning is a challenging task for language models. In addition to exhibiting human-like thinking and reasoning patterns, these models must also be able to store a substantial amount of commonsense knowledge. However, the advent of RALM has made the second requirement less demanding, as retrieval techniques provide language models with access to nonparametric memories. 

FLARE \cite{2} uses StrategyQA, which contains a significant number of yes/no questions from a diverse range of sources. In addition, the authors request that the model provide the exact reasoning process and the final answer that determines the yes/no answer, ensuring that the answer matches the gold answer exactly. The incorporation of in-context samples into the retrieved content, along with training using data from the COPA, HellaSwag, and PIQA datasets, has resulted in the development of LLM-R \cite{164} model that exhibits excellent performance. The fundamental concept of the ITER-RETGEN \cite{165} model employs an iterative methodology to integrate retrievers and language models. The model was trained for the Commonsense Reasoning task using the StrategyQA dataset and achieved its optimal performance at seven iterations. In contrast, KG-BART \cite{166} is designed to prioritize the Commonsense Reasoning task and employs knowledge graphs to enhance its performance in this area. This approach has proven effective in significantly improving the model's ability to complete the Commonsense Reasoning task, with performance approaching that of human beings under certain evaluation metrics.

\subsection{RALM on both NLU and NLG tasks}

\subsubsection{Text Summarization}
Text summarization represents a crucial application of language modelling. In essence, text summarization is the process of generating concise and fluent summaries while maintaining the content and overall meaning of key information. Currently, two distinct types of this task exist: extractive summarization and abstractive summarization.

RA-DIT \cite{26} employs the CNN/DailyMail dataset to refine the language model component of the model, which demonstrates remarkable efficacy in the text summarization task due to the operation of command fine-tuning. In contrast, Self-Mem\cite{17} was trained using the XSum and BigPatent datasets. The authors observed that memory enhancement had a significantly larger effect on BigPatent than XSum. They hypothesize that this discrepancy is due to the inclusion of official patent documents in the BigPatent dataset, which exhibit considerable similarity. The LLM-R \cite{164} model employs an in-context learning approach, integrating RALM, and utilizes the AESLC, AG News, and Gigaword datasets for text summarization training. The results demonstrate that LLM-R significantly outperforms both traditional and dense retrievers in the summarization task. RAMKG \cite{165} extended the iterative training of the RALM architecture to the multilingual domain and employed two multilingual datasets, EcommerceMKP and AcademicMKP, for the training of summarization work, achieving the best results at that time.

\subsubsection{Question Answering}
Question answering also includes generative and extractive forms. It is a common task in NLP that relies on domain-specific knowledge. RALMs can achieve better results than traditional language models by utilizing externally stored knowledge. Common question answering tasks include domain-specific QA, open-domain QA(ODQA), and multi-hop QA. In the medical field, large language models are commonly used, PKG\cite{159} uses the relevant data from the MedMC-QA dataset, using the questions in the training set as input and the medical explanations as background knowledge, and the accuracy of the background knowledge generated by the model as an evaluation metric. HyKGE\cite{160} also targets question answering in the medical field, but uses a knowledge graph-enhanced approach. When targeting ODQA tasks, RAG \cite{19} considers questions and answers as input-output text pairs and trains by minimizing the negative log-likelihood of the answers. In contrast, ICRALM \cite{13} exclusively performs the ODQA task using frozen LMs, which have not been enhanced by pre-training, fine-tuning, or retrieval, as well as the associated knowledge documents. Other models \cite{14} \cite{26} \cite{41} were also trained for the ODQA task. In relation to the multi-hop QA task, FILCO \cite{14} used their proposed filtered retrieval method to filter multiple documents. They validated their approach using the HotpotQA dataset. RR \cite{40}, on the other hand, used the Chain-of-Thought (CoT) approach to address the multi-hop problem. In addition, many other models \cite{2} \cite{42} deal with multi-hop problems.

\subsubsection{Code Generation and Summarization}
Code generation \cite{romera2024mathematical,ye2024reevo} and summarization \cite{nam2024using} differ from ordinary text generation and summarization in terms of the target audience and processing. Code generation and summarization involves computer program code, which may require domain-specific syntactic and semantic understanding, in addition to higher requirements for NLU and NLG capabilities of language models.

REDCODER \cite{73} initially identified potential candidate codes from existing code or abstract databases. The researchers retained 1.1 million unique lines of codes and abstracts as retrieved data through multiple channels. The final evaluation on the CodeXGLUE dataset demonstrated excellent performance across multiple programming languages and evaluation metrics. Another proposed enhancement was put forth by \citet{171}, who utilized private libraries to enhance the quality of code generation. This involved first identifying the most suitable private libraries based on the API documentation through the APIRetriever component, and then utilizing the APIcoder for generation. This approach led to a notable improvement in the accuracy of the generated content.

\citet{75} propose a novel attention-based dynamic graph to complement the source code for static graph representations and design a hybrid message-passing GNN to capture local and global structural information. This approach improves the accuracy of code summarization and ultimately yields superior performance over both mainstream retrievers and generators. The RACE \cite{172} model employ a conventional RALM framework to consolidate code in five programming languages within the MCMD dataset, resulting in a 6 to 38 percent enhancement over all baseline models.

\begin{CJK}{UTF8}{gbsn}
\begin{table*}[]
\caption{Summary of evaluation methods in RALM.}
\label{evaluation}
\renewcommand{\arraystretch}{1.3}
\scalebox{0.68}[0.68]{
\begin{tabular*}{1.46\linewidth}{|l|lllll|l|}
\hline
Reference                                                           & \multicolumn{1}{l|}{\begin{tabular}[c]{@{}l@{}}RAGAS\\ \cite{47}\end{tabular}} & \multicolumn{1}{l|}{\begin{tabular}[c]{@{}l@{}}RGB\\ \cite{128}\end{tabular}} & \multicolumn{1}{l|}{\begin{tabular}[c]{@{}l@{}}CRUD-RAG\\ \cite{129}\end{tabular}} & \multicolumn{1}{l|}{\begin{tabular}[c]{@{}l@{}}ARES\\ \cite{130}\end{tabular}} & \begin{tabular}[c]{@{}l@{}}MIRAGE\\ \cite{174}\end{tabular}                                   & \begin{tabular}[c]{@{}l@{}}RECALL\\ \cite{173}\end{tabular} \\ \hline
Dataset                                                             & \multicolumn{1}{c|}{WikiEval}                                                  & \multicolumn{3}{c|}{LLM-generated}                                                                                                                                                                                                                  & \multicolumn{1}{c|}{\begin{tabular}[c]{@{}c@{}}MMLU-Med\\ MedQA-US\\ MedMCQA\\ PubMedQA\\ BioASQ-Y/N \end{tabular}} & \multicolumn{1}{c|}{\begin{tabular}[c]{@{}c@{}}EventKG\\ UJ\end{tabular}}        \\ \hline
Target                                                              & \multicolumn{5}{c|}{\begin{tabular}[c]{@{}c@{}}Retrieval Quality;\quad Generation Quality\end{tabular}}                                                                                                                                                                                                                                                                                                                                  & Generation Quality                                          \\ \hline
\begin{tabular}[c]{@{}l@{}}Context\\ Relevance\end{tabular}         & \multicolumn{1}{c|}{√}                                                         & \multicolumn{1}{l|}{}                                                         & \multicolumn{1}{l|}{}                                                              & \multicolumn{1}{c|}{√}                                                         &                                                                                               &                                                             \\ \hline
\begin{tabular}[c]{@{}l@{}}Faithfulness\\  \end{tabular}            & \multicolumn{1}{c|}{√}                                                         & \multicolumn{1}{c|}{√}                                                        & \multicolumn{1}{c|}{√}                                                             & \multicolumn{1}{c|}{√}                                                         & \multicolumn{1}{c|}{√}                                                                                            & \multicolumn{1}{c|}{√}                                                          \\ \hline
\begin{tabular}[c]{@{}l@{}}Answer\\ Relevance\end{tabular}          & \multicolumn{1}{c|}{√}                                                         & \multicolumn{1}{l|}{}                                                         & \multicolumn{1}{l|}{}                                                              & \multicolumn{1}{c|}{√}                                                         &                                                                                               &                                                             \\ \hline
\begin{tabular}[c]{@{}l@{}}Noise\\ Robustness\end{tabular}          & \multicolumn{1}{l|}{}                                                          & \multicolumn{1}{c|}{√}                                                        & \multicolumn{1}{l|}{}                                                              & \multicolumn{1}{l|}{}                                                          &                                                                                               &                                                             \\ \hline
\begin{tabular}[c]{@{}l@{}}Information\\ Integration\end{tabular}   & \multicolumn{1}{l|}{}                                                          & \multicolumn{1}{c|}{√}                                                        & \multicolumn{1}{c|}{√}                                                             & \multicolumn{1}{l|}{}                                                          & \multicolumn{1}{c|}{√}                                                                                             &                                                             \\ \hline
\begin{tabular}[c]{@{}l@{}}Negative\\ Rejection\end{tabular}        & \multicolumn{1}{l|}{}                                                          & \multicolumn{1}{c|}{√}                                                        & \multicolumn{1}{l|}{}                                                              & \multicolumn{1}{l|}{}                                                          &                                                                                               &                                                             \\ \hline
\begin{tabular}[c]{@{}l@{}}Counterfactual\\ Robustness\end{tabular} & \multicolumn{1}{l|}{}                                                          & \multicolumn{1}{c|}{√}                                                        & \multicolumn{1}{c|}{√}                                                             & \multicolumn{1}{l|}{}                                                          &                                                                                               & \multicolumn{1}{c|}{√}                                                           \\ \hline
\begin{tabular}[c]{@{}l@{}}Error\\ Correction\end{tabular}          & \multicolumn{1}{l|}{}                                                          & \multicolumn{1}{l|}{}                                                         & \multicolumn{1}{c|}{√}                                                             & \multicolumn{1}{l|}{}                                                          &                                                                                               &                                                             \\ \hline
\begin{tabular}[c]{@{}l@{}}Summarization\\  \end{tabular}           & \multicolumn{1}{l|}{}                                                          & \multicolumn{1}{l|}{}                                                         & \multicolumn{1}{c|}{√}                                                             & \multicolumn{1}{l|}{}                                                          &                                                                                               &                                                             \\ \hline
\end{tabular*}}
\end{table*}
\end{CJK}

\section{Evaluation}
\label{s8}
This section provides a summary of the evaluation approach and benchmarks for RALM. In Sections \ref{s6} and \ref{s7}, we presented some evaluation criteria for large language model tasks as well as baselines. At the time of the initial proposal of the RALM architecture, the majority of researchers employed generalized benchmarks. However, as the RALM architecture evolved, there was a growing number of RALM-specific evaluation methods and baselines proposed. Table \ref{evaluation} demonstrates the details of each evaluation model.

RAGAS \cite{47} employs the WikiEval Dataset to assess the faithfulness, answer relevance, and context relevance of RALMs. Faithfulness is defined as the degree to which responses align with the provided context. Answer relevance refers to the extent to which generated responses address the actual question posed. Context relevance is gauged by the degree to which retrieved context is centralized and devoid of irrelevant information. Additionally, the researchers utilize the prompt gpt-3.5-turbo-16k model to automate the evaluation process. RGB \cite{128} developed a bilingual Chinese and English evaluation system employing three evaluation metrics: accuracy, rejection rate, and error detection rate. These metrics were utilized to assess the noise robustness, negative rejection, information integration, and counterfactual robustness of the data sources, which were articles processed by LM and retrieved through Google's API. CRUD-RAG \cite{129} considers the impact of retrieval components and the construction of external knowledge bases that have not been previously considered by researchers. A dataset was generated using a large model to evaluate the Create, Read, Update, and Delete (summarization) capabilities of RALM through four evaluation metrics: ROUGE, BLEU, bertScore, and RAGQuestEval. In addition, ARES \cite{130} employs datasets generated by the LM, but utilizes a lightweight LM to determine the quality of individual RALM components and utilizes human-labeled data points for prediction-powered inference. The RALM's context is evaluated using the KILT and SuperGLUE benchmarks, with Relevance, Answer Faithfulness, and Answer Relevance being the relevant criteria.

In addition to the general assessment of RALM, there has been some work focusing on the assessment of specific details and domains. RECALL \cite{173} employs the EventKG and UJ datasets to incorporate inaccurate information into its existing data set. It then determines whether RALM is susceptible to being misled by such inaccurate information through two tasks: question answering and text generation. \citet{174} concentrated on the medical domain and proposed MIRAGE, which integrates data from five datasets, including MMLU-Med, to evaluate the zero-shot learning, multi-choice evaluation, retrieval-augmented generation, and question-only retrieval ideation capabilities of medical RALMs. Ultimately, they also discovered the log-linear scaling property and the "lost-in-the-middle" effect in the medical domain.

\begin{figure*}[t]
    \centering
    \includegraphics[width=\textwidth]{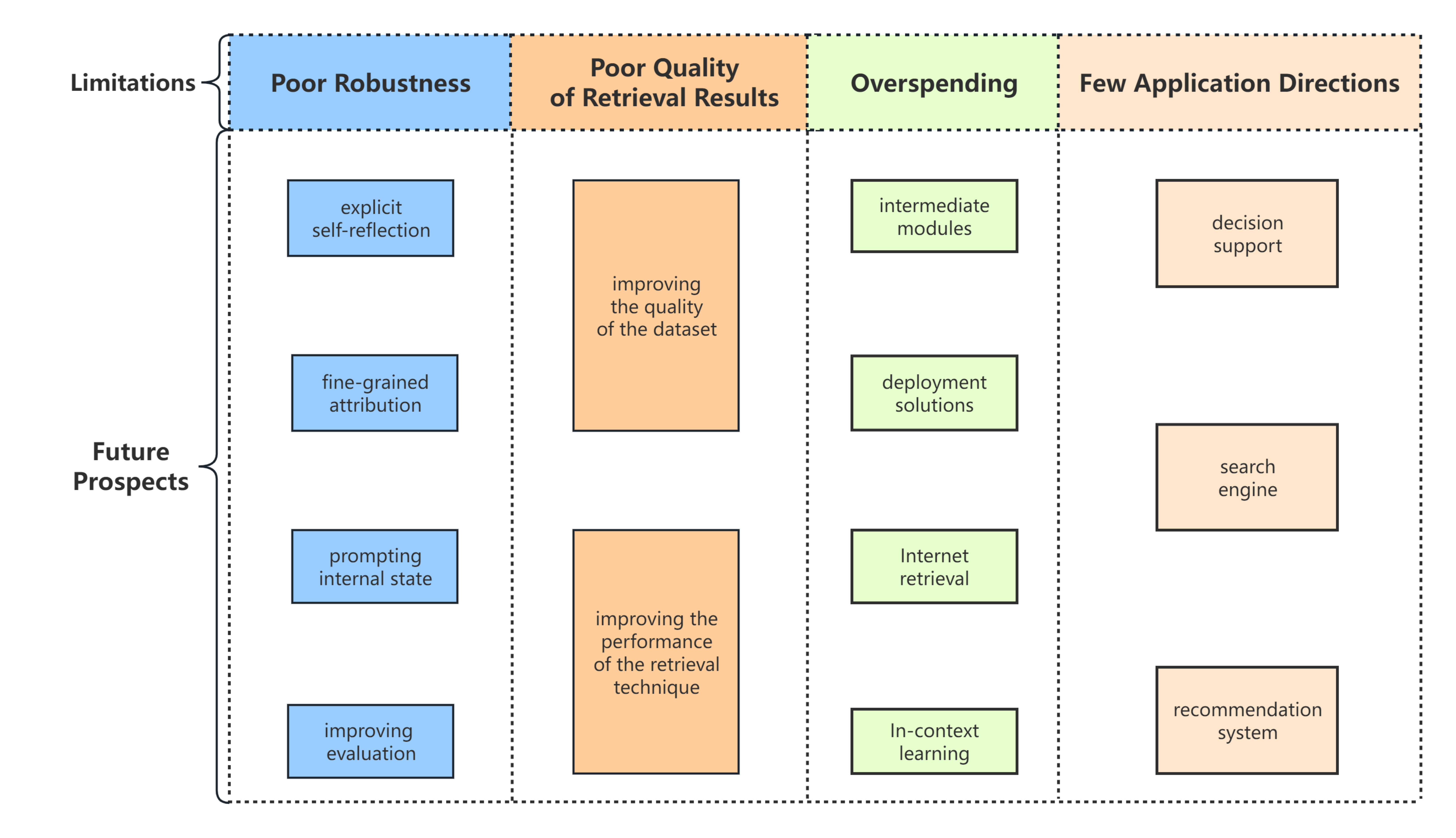}
    \caption{Summary of limitations of current RALM models and future prospects.}
    \label{disscusion}
\end{figure*}

\section{Disscussion}
\label{s9}
This section is devoted to an analysis of the limitations of existing RALM architectures and a description of potential future developments. Figure \ref{disscusion} summarizes the limitations of existing RALMs and our proposed solutions.
\subsection{Limitations}
\label{s9.1}
This section presents a summary and analysis of some of the limitations of existing RALMs.
\subsubsection{Poor Robustness}
Robustness is a crucial aspect to be considered in all systems. RALM systems, despite exhibiting performance benefits in several domains, introduce a multitude of uncertainties to the architecture due to the incorporation of retrieval as a technique. As elucidated by \citet{46}, through exceedingly simple prefix attacks, not only can the relevance and accuracy of RALM output be diminished, but even the retrieval strategy of the retriever can be altered. Consequently, in addition to utilising various retrieval enhancement techniques to enhance the performance of LMs, researchers should also take care to minimise the effects of factually inaccurate data, information that is not relevant to problem solving and even some harmful hints and prefixes on LMs.

\subsubsection{Poor Quality of Retrieval Results}
A significant number of researchers \cite{3} \cite{4} engaged in the endeavour of enhancing retrieval efficacy have asserted that although their proposed models are demonstrably beneficial in optimising the quality of the output, there is as yet no assurance that the retrieval outcomes can be entirely aligned with the LM. Particularly when using the Internet as a retrieval tool, the quality of Internet sources can vary widely, and merging this data without proper consideration can introduce noise or misleading information into the resulting output.

\subsubsection{Overspending}
While existing RALMs \cite{9} \cite{11} \cite{21} can greatly improve the performance of LMs in various domains, some of them require extensive model changes as well as complex pre-training and fine-tuning operations, which greatly increases the time and space overhead and also reduces the scalability of RALMs. In addition, as the scale of retrieval increases, so does the complexity of storing and accessing the data sources. As a result, researchers must weigh the benefits of modifying the model against the costs.

\subsubsection{Few Applications}
Although many RALMs have greatly improved the performance of LMs in various domains, there has not been much improvement from an application perspective, and RALMs are still doing some of the routine work that was done in the early days of LMs, e.g., question answering, summarizing \cite{159} \cite{160} \cite{16}. Although there have been some very interesting application directions recently, such as math teaching \cite{69}, slot filling \cite{81}, etc., this is not enough. A technology always needs to be actually used to fully prove its value, and RALM is no exception.

\subsection{Future Prospects}
This section suggests some possible directions for the future development of RALM, based mainly on the limitations mentioned in Section \ref{s9.1}.

\subsubsection{Improve Robustness}
Some scholars have mentioned possible ways to improve model robustness in the future work section of their papers, such as explicit self-reflection and fine-grained attribution \cite{4}. In contrast to these works, \citet{46} proposed a method called Gradient Guided Prompt Perturbation (GGPP), a way to perturb the RALM, which was experimentally found to be effective in improving the situation by utilizing the SAT probe \cite{175} and activation (ACT) classifier. A method is proposed to detect this perturbation by prompting the internal state of the perturbed RALM. In addition, by proposing and improving the evaluation method of RALM and the related baseline can also help improve the robustness of the model, \citet{128} made a series of evaluation system for RALM by focusing on the robustness.

\subsubsection{Improve Retrieval Quality}
Improving the quality of retrieval can be considered in two parts: improving the quality of the dataset used for retrieval, and improving the performance of the retrieval technique. Nowadays, many data sets are given to LLM to generate relevant content, and since LLM itself has "hallucination", certain means must be adopted to ensure the accuracy of the data, such as using human beings to supervise the refinement \cite{128}. In addition, due to the wide range of information sources on the Internet, it is obviously not enough to rely solely on search engines for screening, so it is necessary to improve the retrieval technology, such as the use of BM25 \cite{27} or TF-IDF \cite{39} algorithms for further re-ranking.

\subsubsection{Weigh Expenses and Benefits}
Reducing the overhead can be considered from three perspectives: first, some plug-and-play intermediate modules can be designed, e.g., CRAG \cite{3}, Selfmem \cite{17}, AI agent \cite{24}, or some deployment solutions, e.g., LangChain, Llama Index, so that there is no need to make targeted improvements for each model. Second, Internet retrieval can be utilized to reduce the overhead of the retriever, but attention needs to be paid to the data relevance mentioned earlier. Finally, In-context learning can be employed to reduce the overhead associated with improving LMs, e.g., ICRALM \cite{13}.

\subsubsection{Expand Applications}
In the contemporary era, the application of LLM has been expanded to encompass a multitude of domains, whereas the application direction of RALM remains relatively limited. To address this limitation, researchers must not only consider the existing application areas of LLM but also leverage the distinctive strengths of RALM, which excels in addressing problems closely related to knowledge and experience. Additionally, they should integrate RALM with other advanced technologies and utilize it to overcome the challenges associated with them. This paper presents several illustrative examples, including decision support, search engine, and recommendation system.

\section{Conclusion}
The integration of RALMs represents a significant advance in the capabilities of NLP systems. This survey has provided an extensive review of RALMs, highlighting their architecture, applications, and the challenges they face. RALMs enhance language models by retrieving and integrating external knowledge, leading to improved performance across a variety of NLP tasks, including translation, dialogue generation, and knowledge graph completion.

Despite their successes, RALMs encounter several limitations. Notably, their robustness against adversarial inputs, the quality of retrieval results, the computational costs associated with their deployment, and a lack of diversity in application domains have been identified as areas requiring further attention. To address these, the research community has proposed several strategies, such as improving the evaluation methods, refining retrieval techniques, and exploring cost-effective solutions that maintain a balance between performance and efficiency.

In the future, the advancement of RALMs will depend on the enhancement of their robustness, the improvement of retrieval quality, and the expansion of their application scope. By incorporating more sophisticated techniques and integrating RALMs with other AI technologies, these models can be leveraged to address an even broader spectrum of challenges. The ongoing research and development in this field are expected to result in more resilient, efficient, and versatile RALMs, thereby pushing the boundaries of what is achievable in NLP and beyond. As RALMs continue to evolve, they hold the promise of enabling AI systems with deeper understanding and more human-like language capabilities, thereby opening up new possibilities in a wide range of fields.

\bibliography{ralm_survey}

\end{document}